\documentclass[default, iicol]{sn-jnl}% Default with double column layout

\usepackage{graphicx}      % include this line if your document contains figures
\usepackage{natbib}        % required for bibliography
\usepackage{bm}
\usepackage{amsmath}
\usepackage{mathtools}
\usepackage{amssymb}
\usepackage{booktabs}
\usepackage{numprint}
\usepackage{caption}
\usepackage{subcaption}
\usepackage{makecell}
\usepackage{algorithm}
\usepackage{xspace}
\usepackage{enumitem}
\usepackage{siunitx}
\usepackage{bm}
\usepackage{amsmath}
\usepackage{xcolor}
\newcommand{\figref}[1]{Fig.~\ref{#1}}

\newcommand{\Rdim}{\ensuremath\mathbb{R}}
\newcommand{\stateScalar}{z}
\newcommand{\stateApprox}{\tilde{\state}}
\newcommand{\stateRec}{\breve{\state}}
\newcommand{\stateSpace}{\mathcal{Z}}

\newcommand{\stateCentered}{\hat{\state}}

\newcommand{\nSamplesTestNum}{100}
\newcommand{\solutionManifold}{\mathcal{S}}

\newcommand{\redIterator}{l}

\newcommand{\disp}{q}
\newcommand{\disps}{\bm{\disp}}
\newcommand{\dispSpace}{\mathcal{Q}}
\newcommand{\stressScalar}{\sigma}
\newcommand{\stress}{\bm{\stressScalar}}
\newcommand{\stressSpace}{\mathcal{G}}
\newcommand{\compTime}{\Delta T}

\newcommand{\redStateDim}{r}
\newcommand{\redStateSpace}{\mathcal{Z}_r}
\newcommand{\redState}{\bar{\state}}
\newcommand{\redStateApprox}{\tilde{\redState}}

\newcommand{\stateDim}{N}
\newcommand{\stateIterator}{n}
\newcommand{\param}{\bm{\mu}}
\newcommand{\paramSpace}{\mathcal{M}}
\newcommand{\paramSet}{\paramSpace_\nSnaps}

\newcommand{\snaps}{\bm{Z}}
\newcommand{\nSnaps}{\kappa}

\newcommand{\redMatrix}{\bm{V}}
\newcommand{\redBasis}{\bm{v}}
\newcommand{\redBasisOpt}{\bm{u}}

\newcommand{\dispDofs}{144636}

\newcommand{\nNodesNum}{48212}
\newcommand{\nodeIterator}{m}

\newcommand{\nElementsNum}{169278}
\newcommand{\muscleDofs}{5}
\newcommand{\nFeat}{n_\text{f}}
\newcommand{\nItem}{n_\text{i}}
\newcommand{\stateDimDisp}{\stateDim_{\text{disp}}}
\newcommand{\stateDimStress}{\stateDim_{\text{stress}}}

\newcommand{\Flow}{\bm{F}}
\newcommand{\FlowApprox}{\tilde{\bm{F}}}
\newcommand{\regression}{\bm{\Phi}}
\newcommand{\reduction}{\bm{\Psi}}
\newcommand{\reconstruction}{\bm{\Psi}^{\dagger}}
\newcommand{\reductionLin}{\bm{V}^T}
\newcommand{\reconstructionLin}{\bm{V}}
\newcommand{\projector}{\bm{P}}

\newcommand{\covariance}{\bm{S}}
\newcommand{\eigenvalue}{\lambda}
\newcommand{\singvalue}{\varsigma}
\newcommand{\singvalues}{\bm{\singvalue}}

\newcommand{\featureSpace}{\mathcal{F}}
\newcommand{\featureMap}{\bm{\Theta}}
\newcommand{\featureDim}{\eta}
\newcommand{\kernelParam}{\gamma}
\newcommand{\kernelBias}{{c}_0}
\newcommand{\kernelDegree}{d}
\newcommand{\kpcaCoef}{a}
\newcommand{\kpcaAlpha}{\varepsilon }

\newcommand{\dataset}{\bm{D}}
\newcommand{\mlInput}{x}
\newcommand{\mlOutput}{y}
\newcommand{\mlInputs}{\bm{\mlInput}}
\newcommand{\mlOutputs}{\bm{\mlOutput}}

\newcommand{\loss}{\mathcal{L}}

\newcommand{\weight}{{w}}
\newcommand{\weights}{\bm{\weight}}
\newcommand{\Weights}{\bm{W}}
\newcommand{\layerIterator}{l}
\newcommand{\neuronIterator}{j}
\newcommand{\activationFunction}{h}

\newcommand{\nUnits}{n_\text{u}}
\newcommand{\nLayers}{n_\text{l}}
\newcommand{\network}{\bm{\Xi}}
\newcommand{\autoencoder}{\network_{\text{ae}}}
\newcommand{\reductionAE}{\reduction_{\text{ae}}}
\newcommand{\reconstructionAE}{\reconstruction_{\text{ae}}}
\newcommand{\seluAlpha}{\alpha}
\newcommand{\seluScale}{\beta}
\newcommand{\vaeMean}{\bm{\nu}}
\newcommand{\vaeVar}{\bm{\delta}}
\newcommand{\vaePrior}{p}
\newcommand{\vaePosterior}{b}
\newcommand{\vaeDis}{\beta}
\newcommand{\vaeNoise}{\bm{\zeta}}

\newcommand{\kernel}{k}
\newcommand{\kernelFunc}{\bm{\kernel}}
\newcommand{\kernelMatrix}{\bm{K}}
\newcommand{\meanFunc}{\bm{m}}
\newcommand{\covMatrix}{\bm{\Sigma}}

\newcommand{\score}{s}
\newcommand{\sRec}{{\score}_{\text{rec}}}
\newcommand{\sRegr}{\ensuremath{\score}_{\text{regr}}}
\newcommand{\sApprox}{\ensuremath{\score}_{\text{appr}}}
\newcommand{\sRecMean}{\ensuremath\hat{\score}_{\text{rec}}}
\newcommand{\sRegrMean}{\ensuremath\hat{\score}_{\text{regr}}}
\newcommand{\sApproxMean}{\ensuremath\hat{\score}_{\text{appr}}}
\newcommand{\eTwo}{\ensuremath e_{\text{2}}}
\newcommand{\eTwoMax}{\ensuremath \eTwo^{\text{max}}}
\newcommand{\eTwoMean}{\ensuremath \eTwo^{\text{mean}}}
\newcommand{\eDist}{\ensuremath e_{\text{dist}}}
\newcommand{\eStress}{\ensuremath e_{\text{stress}}}

\newcommand{\FE}{\textsf{FE}\xspace}

\newcommand{\MOR}{\textsf{MOR}\xspace}
\newcommand{\PCA}{\textsf{PCA}\xspace}
\newcommand{\POD}{\textsf{POD}\xspace}
\newcommand{\KPCA}{\textsf{KPCA}\xspace}

\newcommand{\GP}{\textsf{GP}\xspace}
\newcommand{\GPs}{\textsf{GPs}\xspace}
\newcommand{\NN}{\textsf{NN}\xspace}

\newcommand{\CNNs}{\textsf{CNNs}\xspace}

\renewcommand{\AE}{\textsf{AE}\xspace}
\newcommand{\AEs}{\textsf{AEs}\xspace}
\newcommand{\VAE}{\textsf{VAE}\xspace}
\newcommand{\VAEs}{\textsf{VAEs}\xspace}
\newcommand{\LORE}{\textsf{LRR}\xspace}
\newcommand{\SELU}{\textsf{SELU}\xspace}
\newcommand{\AV}{\textsf{AR/VR}\xspace}

\newcommand{\stateLoc}[1]{#1}%

\renewcommand{\state}{\bm{z}}

\jyear{2022}%

\theoremstyle{thmstyleone}%
\theoremstyle{thmstyletwo}%

\theoremstyle{thmstylethree}%
\begin{document}

\title[Article Title]{Low-dimensional Data-based Surrogate Model of a Continuum-mechanical Musculoskeletal System Based on Non-intrusive Model Order Reduction}

%%=============================================================%%
%% Prefix	-> \pfx{Dr}
%% GivenName	-> \fnm{Joergen W.}
%% Particle	-> \spfx{van der} -> surname prefix
%% FamilyName	-> \sur{Ploeg}
%% Suffix	-> \sfx{IV}
%% NatureName	-> \tanm{Poet Laureate} -> Title after name
%% Degrees	-> \dgr{MSc, PhD}
%% \author*[1,2]{\pfx{Dr} \fnm{Joergen W.} \spfx{van der} \sur{Ploeg} \sfx{IV} \tanm{Poet Laureate} 
%%                 \dgr{MSc, PhD}}\email{iauthor@gmail.com}
%%=============================================================%%

\author[1]{\fnm{Jonas} \sur{Kneifl} }\email{jonas.kneifl@itm.uni-stuttgart.de}

\author[2]{\fnm{David} \sur{Rosin}}\email{rosin@imsb.uni-stuttgart.de}
%\equalcont{These authors contributed equally to this work.}

\author[2]{\fnm{Oliver} \sur{R\"ohrle}}\email{roehrle@simtech.uni-stuttgart.de}
%\equalcont{These authors contributed equally to this work.}

\author*[1]{\fnm{J\"org} \sur{Fehr}}\email{joerg.fehr@itm.uni-stuttgart.de}
%\equalcont{These authors contributed equally to this work.}

\affil[1]{\orgdiv{Institute of Engineering and Computational Mechanics}, \orgname{University of Stuttgart}, \orgaddress{\street{Pfaffenwaldring 9}, \city{Stuttgart}, \postcode{70569}, \stateLoc{Baden-W\"urttemberg}, \country{Germany}}}

\affil[2]{\orgdiv{Institute for Modelling and Simulation of Biomechanical Systems}, \orgname{University of Stuttgart}, \orgaddress{\street{Pfaffenwaldring 5a}, \city{Stuttgart}, \postcode{70569}, \stateLoc{Baden-W\"urttemberg}, \country{Germany}}}
%https://orcid.org/0000-0002-1934-6525 Oliver R\"ohrle
%https://orcid.org/0000-0003-2850-1440 Jörg Fehr
%https://orcid.org/0000-0003-3934-6968 Jonas Kneifl
%https://orcid.org/0000-0002-5154-429X David Rosin
%%==================================%%
%% sample for unstructured abstract %%
%%==================================%%

\abstract{In recent decades, the main focus of computer modeling has been on supporting the design and development of engineering prototyes, but it is now ubiquitous in non-traditional areas such as medical rehabilitation. Conventional modeling approaches like the finite element~(FE) method are computationally costly when dealing with complex models, making them of limited use for purposes like real-time simulation or deployment on low-end hardware, if the model at hand cannot be simplified in a useful manner. Consequently, non-traditional approaches such as surrogate modeling using data-driven model order reduction are used to make complex high-fidelity models more widely available anyway. They often involve a dimensionality reduction step, in which the high-dimensional system state is transformed onto a low-dimensional subspace or manifold, and a regression approach to capture the reduced system behavior. While most publications focus on one dimensionality reduction, such as principal component analysis~(PCA) (linear) or autoencoder (nonlinear), we consider and compare PCA, kernel PCA, autoencoders, as well as variational autoencoders for the approximation of a structural dynamical system. In detail, we demonstrate the benefits of the surrogate modeling approach on a complex FE model of a human upper-arm. We consider both the models deformation and the internal stress as the two main quantities of interest in a FE context. By doing so we are able to create a computationally low cost surrogate model which captures the system behavior with high approximation quality and fast evaluations.
}

%TODO: ORCID einfügen
\keywords{Model Reduction, Machine Learning, Human body models, Kernel PCA, Variational Autoencoder}
\maketitle
\section*{Acknowledgment}
Funded by Deutsche Forschungsgemeinschaft (DFG, German Research Foundation) under Germany's Excellence Strategy - EXC 2075 - 390740016. We acknowledge the support by the Stuttgart Center for Simulation Science (SimTech). 

Furthermore, the authors would like to thank the Ministry of Science, Research and Arts of the Federal State of Baden-W\"urttemberg for the financial support within the InnovationsCampus Future Mobility. 
% add ICM statement
%Acknowledgments of people, grants, funds, etc. should be placed in a separate section on the title page. The names of funding organizations should be written in full.
\section{Introduction}\label{sec1}
% Journal: Archive of Applied Mechanics
%\begin{enumerate}
%	\item Motivation: Real-time application
%	\begin{enumerate}
%		\item AR / VR $\checkmark$
%  		\item computer aided surgery $\checkmark$
%	\end{enumerate}
%	\item structural dynamics not capable
% 	\begin{enumerate}
%		\item separate from multibody $\checkmark$
%   		\item HIL type  $\checkmark$
%	\end{enumerate} 
%	\item surrogate modeling
%  	\begin{enumerate}
%		\item previous research
%  		\item general surrogate modeling approaches $\checkmark$
%		\item other approximation methods, like completely data-based ones or sparse-grid $\checkmark$
%		\item \LORE  $\checkmark$
%  		\begin{enumerate}
%			  \item separate from other \LORE  $\checkmark$
%		\end{enumerate}
%	\end{enumerate} 
%	\item our approach $\checkmark$
% 	\item overview $\checkmark$
%\end{enumerate}
With the growing availability of virtual  and augmented reality (\AV) devices, interest in scientific applications for this type of hardware is also increasing across different research fields, e.g.  immersive data and simulation analysis or computer aided surgery \cite{fleck2022ragrug,arjomandi2022extended}. Placing the user in an environment that allows for direct, intuitive interaction with data and models, having them properly represented in the same space as the user, or providing crucial information directly in a persons field of vision to aid them in the task they are performing, holds great potential for disciplines, which are - so far - primarily experienced through regular screens. However, as comfort of the wearer, and thus weight and size of the device, are primary concerns for the manufacturers, the computational resources an \AV headset can provide are quite limited. % \cite{Hololens2docs}. 

Musculo-skeletal biomechanics simulations to be run on resource-poor systems, or those with realtime performance requirements usually rely on multibody systems using so-called lumped parameter models like Hill-type muscles for the muscle tissue~\cite{hill1938heat}. Lumped parameter models capture the macroscopic behavior of the muscle tissue in an arrangement of discrete elastic, dampening and a contractile elements. Modelling the skeleton as multibody system they can be used for forward simulation up to the scale of full-body models. While very performant and able to exhibit some key characteristics of muscles they only offer a once-dimensional representation~\cite{schmitt2019dynamics}. Realistic muscle geometries like the three-dimensional distribution of stress and deformation or contact mechanics can not be addressed with this type of model. For these purposes a continuum-mechanical material description in combination with the finite-element method (\FE) is well suited for dealing with the large deformations encountered in human soft tissues, like skeletal muscles. These capabilities do, however, come at a considerable computational cost, making them undesirable for any type of time-sensitive simulation or evaluation heavy applications like inverse problem solving musculo-skeletal biomechanics~\cite{rohrle2017two}. 

\paragraph*{Surrogate modelling}
To reduce evaluation times, surrogate modelling techniques can be applied to capture the fidelity of the original \FE model w.r.t. specific aspects, thus decreasing the overall computational cost per evaluation. A common concept for this purpose is the use of response surfaces, i.e. approximating the dependency of an output quantity of interest on the model's inputs by interpolating it based on a number of known states, using a basis function of choice \cite{babaei2016performance}. One such method, which has already been applied successfully to continuum-mechanical data, is a sparse grid approach in combination with B-spline interpolation \cite{valentin2018gradient}. The name refers to building the support for the response surface as sparse as possible, to maintain the methods usefulness for either particularly expensive models or models with high input dimensionality. Another technique involves a coarse model, with lower fidelity, to explore the parameter space it shares with the original, fine-grained  or high-fidelity model, which is then only evaluated for states identified as useful or interesting using the surrogate \cite{jansson2005optimization}. Variants of this type of approach that have been popularized with the more wide-spread use of machine learning over the last decade, attempt to bypass the full model entirely by learning the difference between the surrogate and the original model which can be called discrepancy learning~\cite{Gupta2021, ebers2022discrepancy}. Intuitive examples are deep-learning-based smart upscaling techniques to increase visual fidelity, e.g. in computer graphics \cite{xiao2020neural}. A similar approach does however also exist for simulations . %or methods like proper orthogonal decomposition, aimed at reducing the dimensionality of the system of equations describing the model during the numerical solution process (cite Box\&Wilson, and Mylena).

\paragraph*{Data-based Model Order Reduction}
In addition to classical machine learning, the field of \emph{model order reduction}~(\MOR) provides a broad library of methods for the efficient approximation of high-dimensional systems.
Besides classical linear techniques like modal or Krylov reduction and parametrized model reduction, as well as classical nonlinear techniques such as collocation-based methods or optimized global cubature methods~\cite{An2008,FarhatEtAl14}, data-driven approaches have became a popular solution and close the gap to machine learning. Their benefits include their capability to handle black- and gray-box models and their ability to capture nonlinear behavior. Furthermore, the disadvantage of having to manipulate simulation code is avoided. Unfortunately, this generally requires a costly offline phase and physical behavior can often no longer be guaranteed.

A concept, which many of the data-driven \MOR approaches share, is the idea of transforming a high-dimensional state onto a low-dimensional space~(\emph{reduction}) where the relationship between input variables and a low-dimensional representation of the state is learned~(\emph{regression}). 
Many publications in this context are focused on the use of specific choice of reduction and regression algorithms but obviously there are endless possible combinations. 
To account for all of them and to understand this idea as a general and modular framework, we will refer to this type of method generally speaking as \emph{low-rank regression} (\LORE). 
An example for this can be found in~\cite{HesthavenUbbiali2018}, where \emph{proper orthogonal decomposition}~(\POD) is combined with a neural network to extract a low-dimensional subspace from data first and predict the reduced dynamics of a parametrized partial differential equation second. The authors refer to this certain approach as \POD-\NN. 

In the field of structural dynamics, crash simulation models in particular have been the objective of previous \LORE research~\cite{GuennecEtAl18, KneiflGrunertFehr21, KneiflHayFehr2021, KapsCzechDuddeck2022}. In these publications, the quantity of interest has been the displacement of the considered models and the results have proven that it is possible to approximate them well using linear reduction techniques like the~\POD or the CUR decomposition~\cite{GuennecEtAl18}. Other physical quantities such as stress, which proved more difficult to capture in our research, have been neglected so far or were only calculated in a post-processing step using standard FEM tools as in~\cite{Fresca2022}. 
Consequently, with this work we apply the \LORE framework to both displacement and stress in order to develop more comprehensive surrogate models. 
Furthermore, we show that this kind of surrogate model is well-applicable for human body models which are rarely considered in data-based \MOR of structural dynamical problems.

When linear reduction techniques reach their limits regarding approximation quality, nonlinear techniques can provide better dimensionality reduction capabilities. In \cite{SalvadorDedeManzoni2021}, kernel proper orthogonal decomposition is used as nonlinear extension of \POD in combination with neural networks to approximate parametrized partial differential equations. Another popular nonlinear dimensionality reduction approach are neural networks in the form of \emph{autoencoders} (\AE)~\cite{Kramer1991}. They are composed of a decoder that maps the high-dimensional data onto a low-dimensional representation and an encoder which follows the task to reconstruct the full data from this reduced representation. 
In \cite{FrescaDedeManzoni2021}, for example, deep convolutional autoencoders are used along with feed-forward neural networks as surrogate models in cardiac electrophysiology. Other applications of convolutional autoencoders in the context of \MOR can be found in \cite{GonzalezBalajewicz2018, LeeCarlberg2020}. Convolutional neural networks are a popular choice for the task of reduction as they possess a lower number of trainable parameters compared to fully connected neural networks. Moreover, for structured, grid-like data, they are able to detect local patterns with filters. Unfortunately, they are not suited for finite element models which in general are not spatially discretized as grid. 
This can be remedied by using graph convolutions~\cite{DefferrardBressonVandergheynst2016}, which are compared to a fully connected neural network in the context of \MOR in~\cite{GruberEtAl2022}. Graph convolutional autoencoders have been applied in the field of computer vision in combination with mesh sampling operations to reconstruct 3D faces \cite{RanjanEtAl2018} from a low-dimensional space. Moreover, many combinations of different reduction methods are possible: autoencoders which complement a linear subspace~\cite{ShenEtAl2021}, a series of linear reduction methods to reduce the system to a certain degree, and subsequent autoencoders to reduce it even further, see, for example,~\cite{ContiEtAl2022}, etc. 
% However, as we want to compare the core methods themselves in this paper no combinations are considered.

In contrast to the aforementioned methods which try to approximate the reduced system using a black-box model, other approaches aim to discover low-dimensional governing equations. In \cite{Champion2019} an autoencoder is used to find a low-dimensional space in which the governing equations of the system are identified using sparse regression on a library of suitable function candidates. This approach can also be extended to discover hidden variables by adding time-delay embedding, see~\cite{Bakarji2022}. In the context of structural dynamics, this kind of approach was used to discover low-dimensional equations of a beam, which can then be used for continuation~\cite{ContiEtAl2022}.

\paragraph*{Scope and Overview}
In this work, we use the \LORE framework to approximate the behavior of a human arm, i.e., its motion in form of node displacements and resulting internal stress. Thereby, we compare
multiple reduction techniques, namely, the frequently used \emph{principal component analysis}~(\PCA), its nonlinear extension \emph{kernel principal component analysis}~(\KPCA) as well as fully-connected classical \emph{autoencoders} (\AE) and \emph{variational autoencoders}~(\VAE) as other nonlinear alternatives. 
For this purpose, reference data is obtained from a high-fidelity \FE model based on different muscle activation levels. With this data the reduction algorithms are fitted and used to create a low-dimensional representation of the data. The relation between muscle activation and reduced coordinates is approximated using a \emph{Gaussian process}~(\GP) regression algorithm. For the approximation of the arm's full behavior, the \GP's predictions are transformed back  into original space. 

The theoretical background can be found in Section~\ref{sec:methods}, the resulting algorithm and its implementation in Section~\ref{sec:implementation}, while the results for the arm model are presented and discussed in Section~\ref{sec:results} followed by the conclusion in Section~\ref{sec:conclusion}.

The contributions of our work can be summarized as:
\begin{enumerate}
	\item we generate real-time capable, yet accurate holistic surrogate models for a complex human arm model based on a 5-dimensional input 
	\item we apply the \LORE framework to data with nonlinear structures such as stress
	\item we introduce \KPCA as a nonlinear alternative to \PCA in \LORE
	\item we compare \PCA, \KPCA, \AE, and \VAE regarding their reconstruction and generalization qualities
\end{enumerate}

%%%%%%%%%%%%%%%%%%%%%%%%%%%%%%%%%%%%%%%%%%%%%%%%%%%%%%%%%%%
\section{Methods}\label{sec:methods}
The underlying approximation method used in this paper relies on a coordinate transformation to find a low-dimensional description of the system and on a regression algorithm to learn the behavior of the system. Although we start with introducing the specific human upper-arm model, the used methodology generalizes well to other problems and is explained in general terms. 

\subsection{Model}
Muscles are the driving organs, i.e. the actuators, of the human body. Together with soft tissues such as tendons, they are essential for the proper interaction and functioning of the human organism. For example, muscle activation has a great impact in car crashes, see~\cite{Osth2022}. It can stiffen extremities, reduce the load on certain parts of the body, and has a huge impact in the likelihood of injuries. 

\begin{figure}
	\centering
	\includegraphics[trim=35cm 15cm 30cm 11cm, clip, width=120pt]{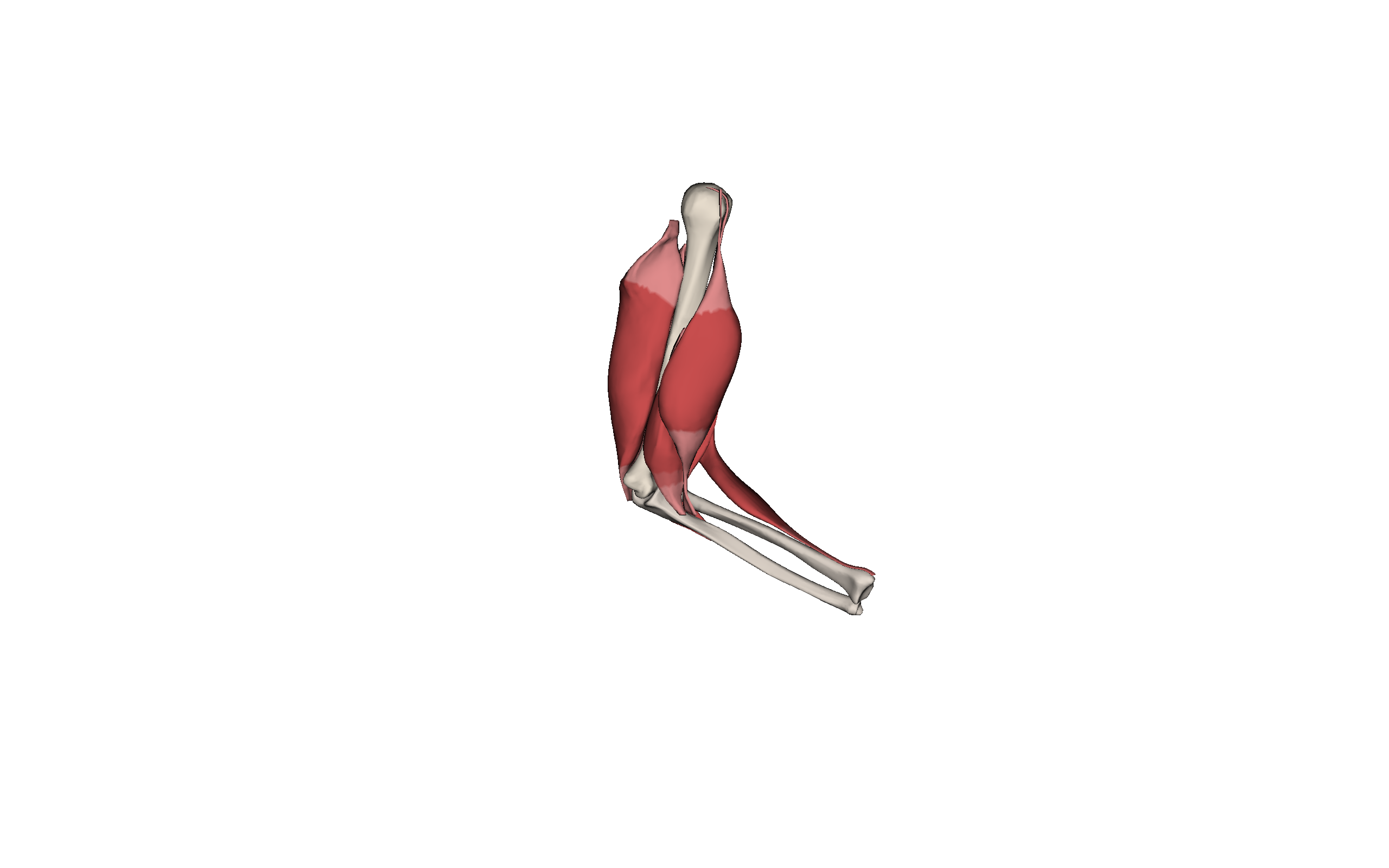}
	\caption{Human upper-arm model including five muscles actuating the elbow joint.}
	\label{fig: arm}
\end{figure}

The considered human upper-arm model consists of the radius and ulna bones for the lower arm and the humerus for the upper-arm, see \figref{fig: arm}. 
The elbow joint connecting them is modelled as a simple hinge joint. It contains five muscle actuating this joint, two extensors and three flexors. Namely these are the \textit{m.~triceps~brachii} and the \textit{m.~anconeus}, as well as the \textit{m.~biceps~brachii}, the \textit{m.~brachialis} and the \textit{m.~brachioradialis}. The geometry of all these components is modelled after the Visible Human Male dataset \cite{ackerman1998visible}. In \cite{rohrle2017two}, a previous iteration of this upper limb model is described and details regarding the constitutive and material modelling are pointed out. Since the model used here heavily relates to the aforementioned one, we will primarily describe parts of the model that differ from the previous version, and those which are most relevant to this work.
The upper head of the humerus bone, normally connected to the shoulder, is fixed in an upright position, making the elbow joint the only mechanical degree of freedom in the model.\\
Input to the model consists of a vector of activation parameters $\param \in \paramSpace \subseteq \Rdim^{\muscleDofs}$. Each activation $\mu_j \in [0,1]$ with $j=1,...,5$ is associated with one muscle in the model, representing the percentage of the maximum possible active stress currently output by the muscle.\\
The overall stress in the $j$-th muscle can be written as a sum of isotropic and anisotropic contributions $\stress_\text{iso}$ and $\stress_\text{aniso}$, of which the latter can be subdivided further into active and passive stresses $\stress_\text{active}$ and $\stress_\text{passive}$:
\begin{flalign*}
    \stress = \stress_\text{iso} + (\stress_\text{passive} + \mu_j\gamma_M\stress_\text{active})(1-\gamma_{ST}).
\end{flalign*}
 The variables $\gamma_M, \ \gamma_{ST} \in [0,1]$ are scale factors describing the tissue, with $\gamma_M,\  \gamma_{ST}>0$ being muscle, $\gamma_M =0, \ 0<\gamma_{ST}\le 1$ tendon, and $\gamma_M=\gamma_{ST}=0$ soft tissues, e.g. fat. %Active stress output corresponds to specific fiber lengths according to the so-called force-length-relation, which is here represented by the optimal fiber length $\lambda_\text{opt}$, for which maximum force output is achieved. As such, this is a type of anisotropic stress, since it depends on the orientation of the muscle fibers.
The active stress is defined as:
\begin{flalign*}
    &\stress_\text{active} = \\ &\begin{cases}
   \frac{\sigma_\text{max}}{\lambda_f^2}\exp{\left(- \left\vert\frac{\lambda_f/\lambda_f^\text{opt}-1}{\Delta \Gamma_\text{asc}}\right\vert^{\nu_\text{asc}}\right)}\mathbf{M}\ \text{for}\ \lambda_f\leq\lambda_f^\text{opt}\\
    \frac{\sigma_\text{max}}{\lambda_f^2}\exp{\left(-\left\vert\frac{\lambda_f/\lambda_f^\text{opt}-1}{\Delta \Gamma_\text{desc}}\right\vert^{\nu_\text{desc}}\right)}\mathbf{M}\ \text{for}\ \lambda_f>\lambda_f^\text{opt}.
    \end{cases}
\end{flalign*}
In this context, $\sigma_\text{max}$ is the maximum possible active stress for the $j$-th muscle. This occurs at the optimal fibre length $\lambda_f^\text{opt}$. More information on this, as well as all other material parameters of the model, can be found in the appendix in Table \ref{tab:params}.

The steepness of the force-length-relation of the muscle fibre is given by $\Gamma_\text{asc}$ and $\nu_\text{asc}$ for the ascending arm of the relation, and $\Gamma_\text{desc}$ and $\nu_\text{desc}$ for the descending arm. $\lambda_f$ is the muscle fibre stretch in the current configuration. $\mathbf{M}=\mathbf{a}_0\otimes\mathbf{a}_0$ is defined as by the fibre orientation in the reference configuration $\mathbf{a}_0$, with $\otimes$ being the dyadic product.

Furthermore the passive anisotropic stress can be written as
\begin{align*}
    \stress_\text{passive} =\ \begin{cases}
    \ \frac{c_3}{\lambda_f^2}\left(\lambda_f^{c_4} - 1\right)\mathbf{M} &\text{for}\ \lambda_f\geq1\\
    \ 0 &\text{else},
    \end{cases}
\end{align*}
where $c_3$ and $c_4$ are material parameters.
%The activations $mu_i$ are then applied in the active stress as a scale factor:\\
%\begin{flalign}
%    \mathbf{S}_\text{aniso} = \mathbf{S}_\text{passive} + \alpha\gamma_M\mathbf{S}_\text{active})(1-\gamma_{ST})
%\end{flalign}
On top of these anisotropic contributions, the muscles also exhibit some isotropic behavior which is given by:\\
\begin{flalign*}
    \stress_\text{iso} = (B_1\mathbf{I}+B_2\mathbf{C}+B_3\mathbf{C}^{-1})\\+k(\rm det\mathbf{F}-1)I_3^{1/2}\mathbf{C}^{-1}
\end{flalign*}
with
\begin{flalign*}
    B_1 &= 2c_1I_3^{-1/3}+2c_2I_3^{-2/3}\mathbf{C}^{-1},\\
    B_2 &= -2c_2I_3^{-2/3},\ \text{and}\\
    B_3 &= -2/3c_1I_3^{-1/3}I_1-4/3c_2I_3^{-2/3}I_2.
\end{flalign*}
Again, $c_1$ and $c_2$ are material parameters. The invariants $I_1, I_2, I_3, I_4$ can be derived from the Cauchy-Green stress tensor $\mathbf{C}$:
\begin{flalign*}
I_1 &= \rm tr\mathbf{C}\\
I_2 &= \rm tr(cof\mathbf{C})\\
I_3 &= \rm det\mathbf{C}\\
I_4 &= \lambda_f^2 =\mathbf{a}\otimes\mathbf{a} = \rm tr(\mathbf{MC}).
\end{flalign*}
%Model geometry taken form the visible human male data-set of a right arm. 
%Shoulder fixed with humerus in an upright position.
%mechanical degree of freedom given by the elbow joint, which is modelled as a hinge joint.
%the  model contains five  muscles, two extensors, three flexors, namely: the m. triceps brachii and m. anconeus, as well as the m. biceps brachii, m. brachialis, m. brachioradialis.
%The model takes five activation parameters as input. One for each muscle. The activation corresponds to percent of maximum contraction of the muscle fibres. According to the force-length-relation each fibre length corresponds to a specific force output.
%Also contact mechanics maybe?
%Formulation of the Stress seems appropriate to illustrate the fibre lenght thingy.
%high-fidelity \emph{black-box} model~$\Flow:\paramSpace \to \stateSpace \subseteq \Rdim^{\stateDim}$
%with muscle activation~$\param \in \paramSpace \subseteq \Rdim^{\muscleDofs}$ and system (output) state vector~$\state \in \stateSpace$ which reflects the~$\nFeat$ features of the model's~$\nItem$ items in a single vector so that~$\stateDim=\nItem\nFeat$.

For the purposes of the approximation methods applied in this work the model described above serves as high-fidelity model and is mathematically defined as~$\Flow:\paramSpace \to \stateSpace \subseteq \Rdim^{\stateDim}$. It possess unknown since not accessible internal dynamics. The  quantities of interest in this work, on the contrary, are accessible as simulation results. They include the motion of the arm in form of its (node-wise) vectorized displacements $\disps \in \dispSpace \subseteq \Rdim^{\stateDimDisp}$ with
\begin{align*}
	\disps = \begin{bmatrix}
		\disps_1 \\
		\disps_2 \\
		\vdots \\
		\disps_{\nItem}
	\end{bmatrix}
\end{align*}
where~
$\disps_\nodeIterator=\begin{bmatrix}
	\disp_{m,x} & \disp_{m,y} & \disp_{m,z}
\end{bmatrix}^T$
represents the $x$,- $y$-, and $z$-displacement of the $m$-th node. The simulation results furthermore contain the (element-wise) stress load of the muscle tissue. For the latter, we consider the equivalent stress values~$\stress \in \stressSpace \subseteq \Rdim^{\stateDimStress}$ according to von Mises 
% ~$\stress=f(\disps) \in \stressSpace \subseteq \Rdim^{\stateDimStress}$ 
\begin{align*}
	\stress^2=\dfrac{1}{{2}}
		({(\stress_x-\stress_y)^2 + (\stress_y - \stress_z)^2 + (\stress_z-\stress_x)^2}) \\
		 {+ 3(\stress_{xy}^2  + \stress_{yz}^2  + \stress_{zx}^2)}
\end{align*}
as it offers advantage to combine all stress components into a single value.
The dimension of the displacements~$\stateDimDisp=\nNodesNum\cdot3=\dispDofs$ corresponds to the product of the number of nodes~$\nItem=\nNodesNum$ and features per node~$\nFeat=3$, while the dimension of the von Mises stress simply equals the number of elements~$\stateDimStress=\nElementsNum$.
In the following explanations, we will not distinguish whether the displacements~$\disps$ or the stress~$\stress$ is the quantity of interest as it makes no difference for the underlying approach. Correspondingly, we introduce the variable~$\state \in \stateSpace \subseteq \Rdim^\stateDim$ as general state vector, which can describe an arbitrary physical quantity.

\subsection{Problem Definition}
Generally speaking, we pursue the goal of creating an efficient but accurate surrogate model~$\FlowApprox$ for a computationally expensive high-fidelity model~$\Flow$. A major problem encountered in surrogate modeling of sophisticated models is their sheer dimensionality. In the case of finite element models, this high dimensionality arises from the underlying modeling method and not from the necessity to describe the system with so many degrees of freedom.  
Fortunately, this often results in high-dimensional systems that can be expressed with only a few intrinsic coordinates, instead of thousands or millions, by a suitable coordinate transformation. 

Consequently, our surrogate modeling approach relies on two fundamental steps. The first step involves the identification of a suitable coordinate transformation~$\reduction: \stateSpace \to \redStateSpace$ mapping the system state~$\state$ onto its reduced low-dimensional equivalent~$\redState \in \redStateSpace \subseteq \Rdim^\redStateDim$ where~$r \ll N$. A reverse coordinate transformation from reduced to full space~$\reconstruction: \redStateSpace \to \stateSpace$ ensures physical interpretability of the results. 
% Please note that~$\reconstruction$ is in general not the true inverse mapping of~$\reduction$ and the notation only indicates that it represents the reverse mapping.

\begin{figure}[]
	\centering
	\includegraphics[]{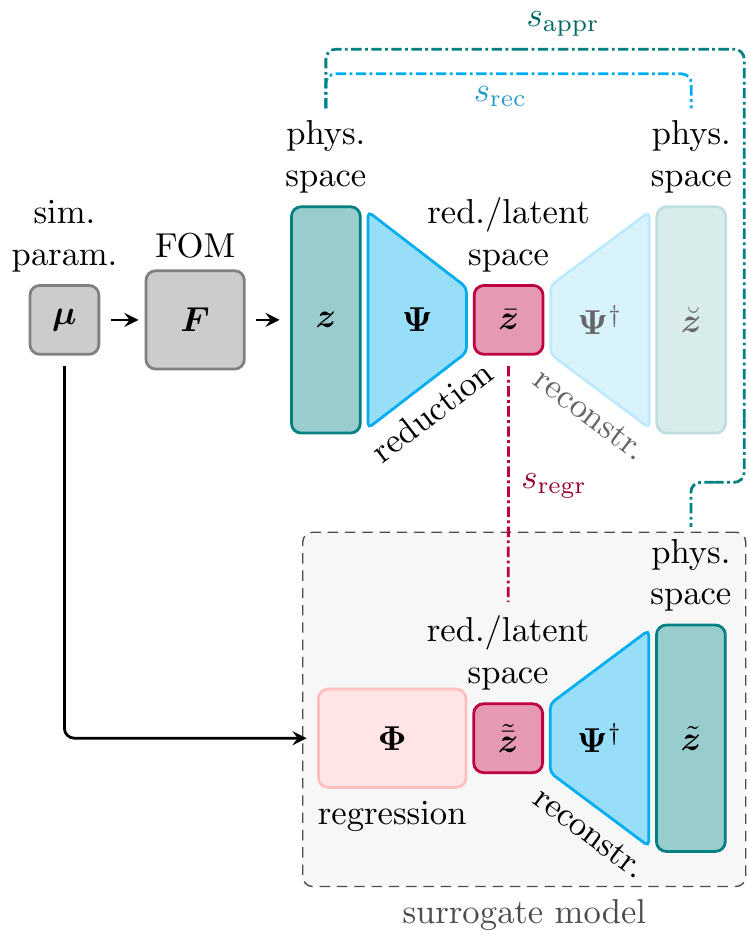}
	\caption{Low-rank regression architecture. Offline stage: evaluate FOM~$\Flow(\param)$, fit~$\reduction$,~$\reconstruction$, and~$\regression$. Online stage: evaluate surrogate model~$\reconstruction \circ \regression(\param)$. Validation: compare quantities~$\sRec$,~$\sRegr$, and~$\sApprox$
	}
	\label{fig:low-rank regression}
\end{figure}

The second step involves the approximation of the system behavior in reduced space, i.e., of the reduced coordinates~$\redState$. Therefore, we seek a function~$\regression: \paramSpace \to \redStateSpace$ that approximates the relationship~$\redState \approx \redStateApprox = \regression(\param)$ between quantities from the parameter space~$\paramSpace$ and reduced system behavior. The abstract workflow can be seen in \figref{fig:low-rank regression}. 
It is important to keep in mind that besides the approximation itself, the surrogate's purpose is to be executable under time- and resource constraints. Therefore, the computation time of the surrogate~$\compTime_{\FlowApprox}$, which is used as a general indicator of the computational effort required, must be significantly lower, than that of the original model~$\compTime_{\Flow}$.  

Taking all mentioned considerations into account, the overall goal can be formulated in a mathematical sense resulting in an optimization problem
\begin{align}
	\min_{\reduction, \reconstruction, \regression} \quad & \underset{\substack{\\\param\in\paramSpace}}{\sum} \quad	
	\loss(\state(\cdot),\stateApprox(\cdot))	
	\label{eq: problem definition} \\
	\text{s.t.} \quad		&\stateApprox(\param) =\reconstruction(\redStateApprox),  \nonumber \\
	 								&{\redStateApprox}(\param) = \regression(\param), \nonumber	\\
									&{\redState}(\param) = \reduction(\state(\param)), \nonumber	\\
									&{\state}(\param) = \Flow(\param), \nonumber	\\
									& \compTime_{\FlowApprox} \ll \compTime_{\Flow} \nonumber.
\end{align}
In this context, the function~$\loss$ is some measure of approximation quality comparing the results from the surrogate~$\stateApprox$ with the reference results~$\state$. The optimization problem \eqref{eq: problem definition} is minimized over~$\reduction$ and~$\reconstruction$ to find the best low-dimensional representation of the system and over~$\regression$ to find the best approximation of the reduced system coordinates. Furthermore, the parameters~$\param \in \paramSpace$ are taken into account to guarantee a satisfying approximation in the complete considered parameter space~$\paramSpace$.

In the following, ~$\reduction$ will be called \emph{reduction mapping} and~$\reconstruction$ accordingly \emph{reconstruction mapping}. Furthermore, an approximation of a quantity is always indicated by a tilde as in~$\stateApprox$, whereas a reduced quantity is represented by a bar, e.g.,~$\redState$, as listed in Table~\ref{tab: most important variables}, where the most important variables of this work are summarized. Parameter dependencies $\state(\param)$ are only mentioned where it helps the comprehensibility and neglected otherwise.
\begin{table}
	\centering
	\caption{Definitions of the most important variables. The goal of this work is to approximate $\state$ by $\stateApprox$ with suitable surrogate models.}
	\label{tab: most important variables}
	% \npdecimalsign{.}
	% \nprounddigits{3}
	\begin{tabular}[c]{c l}
		\toprule
		variable& description\\ 
		\midrule
			$\state$ & reference state\\
			$\redState$ & reduced state\\
			$\stateRec$ & reconstructed state\\
			$\stateApprox$ & approximated state\\
			\\
			$\reduction$ & reduction mapping\\
			$\reconstruction$ & reconstruction mapping\\
			$\regression$ & regression\\
			\\
			$\disps$ & node displacement\\
			$\stress$ & von Mises stress\\
%		\bottomrule
	\end{tabular} 
	\npnoround
\end{table}

%%%
\subsection{Dimensionality Reduction}
\label{sec:model reduction}
For an optimal low-dimensional representation of the system states~$\redState=\reduction(\state)$ a coordinate transformation (reduction mapping)~$\reduction: \stateSpace \to \redStateSpace$ and its reverse transformation (reconstruction mapping~$\reconstruction: \redStateSpace \to \stateSpace$ are sought.
Their function composition~$\stateRec=\reconstruction \circ \reduction(\state)$ yields the reconstructed state.

There exist numerous possibilities to derive the reduction and reconstruction mapping in the field of model order reduction~\cite{GuoHesthaven2019, BennerEtAl2021}.
In the light of data-based approaches, snapshot-based methods are a popular choice. They assemble~$\nSnaps$ high-fidelity simulation samples, so-called snapshots, in the \emph{snapshot matrix}
\begin{align}
	\snaps=
	\begin{bmatrix}
		\state_{1} \
		\state_{2}\
		...\
		\state_{\nSnaps}\
	\end{bmatrix}\in\Rdim^{\stateDim \times \nSnaps}
	\label{eq: snapshot matrix}
\end{align} 
where~$\state_{{k}}=\state(\param_{k})$.
The individual snapshots rely on the discrete and finite parameter set~$\paramSet = \{\param_1,...,\param_\nSnaps\} \subseteq \paramSpace$. As long as there are enough samples in the resulting subspace~${\solutionManifold_{\paramSet}=\text{span}\{\snaps\}}$, it is supposed that this subspace approximates the discrete solution manifold~$\solutionManifold_{\paramSpace}=\{\state(\param) \ \vert \ \param\in\paramSpace\}$. This, in turns, ensures that if $\reduction$ and $\reconstruction$ are found using $\solutionManifold_{\paramSet}$ they can be applied to~$\solutionManifold_{\paramSpace}$ as well.

A widespread methodology to find the reduced coordinates are linear reduced-basis methods, see~\cite{QuarteroniManzoniNegri16}. They rely on the mathematical principle of projection. In case of a Galerkin projection with an orthogonal projector~$\projector$, the state is reconstructed by
\begin{align*}
	\state\approx\stateRec=\projector\state=\reconstructionLin\reductionLin\state=\reconstructionLin\redState .
\end{align*}
In this context, the matrix~$\reconstructionLin=\begin{bmatrix}
	\redBasis_1 & ...&\redBasis_\redStateDim
\end{bmatrix}$ is known as reduction matrix and consists of the \emph{reduced basis vectors}~$\{\redBasis_1,...,\redBasis_\redStateDim\}\subseteq\stateSpace$.  That means the full state is approximated by a
linear combination~$\state\approx\sum_{\redIterator=1}^{r}\bar{z}^{(l)}\ \redBasis_l$ of the reduced basis vectors with the \emph{reduced states}~$\redState = \begin{bmatrix} \bar{z}^{(1)} & ... & \bar{z}^{(\redStateDim)} \end{bmatrix}^T$ as coefficients. 
Using such methods can result in a poor reconstruction or a relatively large number of required modes $\redStateDim$ due to the restriction to describe the state as a linear combination.
Hence, we present besides principal component analysis as a representative of linear projection-based reduction methods, nonlinear alternatives in the form of kernel principal component analysis as well as classical and variational autoencoders.

\subsubsection{Principal Component Analysis}
Principal component analysis, which is also known as proper orthogonal decomposition~\cite{Volkwein13}, works with centered data~$\hat{\snaps}
\hspace{-.4mm}=\hspace{-.4mm}
\begin{bmatrix}
	\hat{\state}_{1} \
	\hat{\state}_{2}\
	...\
	\hat{\state}_{\nSnaps}
\end{bmatrix}$. 
That means the mean~$\state^{\text{mean}}=\frac{1}{\nSnaps}\sum_{n=1}^{\nSnaps}{\state}_n$ is subtracted samplewise~$\hat{\state}=\state-\state^{\text{mean}}$ from the data.
The purpose of~\PCA is to find orthonormal basis vectors, also referred to as principal components ~$\{\redBasis_\redIterator\}_{\redIterator=1}^\redStateDim$ for which the maintained variance under projection
% \begin{align}
% 	\variance_\redIterator=
% 	\frac{1}{\nSnaps}
% 	\sum_{\stateIterator=1}^{\nSnaps}
% 		(\redBasis_\redIterator^T\stateCentered_\stateIterator)^2
% 	&= \frac{1}{\nSnaps} 
% 		\redBasis_\redIterator^T 
% 		\sum_{\stateIterator=1}^\nSnaps \stateCentered_\stateIterator\stateCentered_\stateIterator^T \redBasis_\redIterator
% 		= \redBasis_\redIterator^T \covariance \redBasis_\redIterator
% 	\label{eq:maintained variance}
% \end{align} 
\begin{align*}
	\frac{1}{\nSnaps}
	\sum_{\redIterator=1}^{\redStateDim}
	\sum_{\stateIterator=1}^{\nSnaps}
		(\redBasis_\redIterator^T\stateCentered_\stateIterator)^2
	&= \frac{1}{\nSnaps} 
	\sum_{\redIterator=1}^{\redStateDim}
		\redBasis_\redIterator^T 
		\sum_{\stateIterator=1}^\nSnaps \stateCentered_\stateIterator\stateCentered_\stateIterator^T \redBasis_\redIterator\\
	&=\frac{1}{\nSnaps} 
	\sum_{\redIterator=1}^{\redStateDim}
		\redBasis_\redIterator^T 
		\hat{\snaps}\hat{\snaps}^T \redBasis_\redIterator
	\label{eq:maintained variance}
		% &= \sum_{\redIterator=1}^{\redStateDim}\redBasis_\redIterator^T \covariance \redBasis_\redIterator
	% \reductionLin\sum_{\stateIterator=1}^\nSnaps \stateCentered_\stateIterator\stateCentered_\stateIterator^T \reconstructionLin \\
	% \frac{1}{\nSnaps}\reductionLin\sum_{\stateIterator=1}^\nSnaps \stateCentered_\stateIterator\stateCentered_\stateIterator^T \reconstructionLin 
	% = \reductionLin \covariance \reconstructionLin
\end{align*}
remains maximal, see~\cite{Hotelling1933}.

Accordingly, the reduced basis vectors solve the optimization problem
\begin{eqnarray}
	\{\redBasis_1,...,\redBasis_\redStateDim\}=\arg\max_{\{\redBasisOpt_1,...,\redBasisOpt_\redStateDim\}} & \sum_{\redIterator=1}^{\redStateDim}\redBasisOpt_\redIterator^T \covariance \redBasisOpt_\redIterator \label{eq: variance opt problem} \\
	\text{s.t.} \quad &\redBasisOpt_k^T\redBasisOpt_\stateIterator=\delta_{k\stateIterator} \nonumber
\end{eqnarray}
where~$\covariance=\frac{1}{\nSnaps}\hat{\snaps}\hat{\snaps}^T\in\Rdim^{\stateDim \times \stateDim}$ represents the covariance matrix and~$\delta_{k\stateIterator}$ the Kronecker delta.
One possibility to determine the basis vectors is to solve an eigenvalue problem of the covariance matrix
\begin{align*}
	\covariance\redBasis_\redIterator\ = \eigenvalue_\redIterator \redBasis_\redIterator.
\end{align*}
The reduced basis vectors then result in the~$\redStateDim$~most dominant eigenvectors, i.e., those with the largest associated eigenvalue~${\eigenvalue}$. Usually, the eigenvectors are ordered in descending order of their eigenvalues so that $\eigenvalue_1\geq\eigenvalue_2\geq...\geq\eigenvalue_\redStateDim>\eigenvalue_{\redStateDim+1}\geq...\geq\eigenvalue_\nSnaps$, where we suppose that the $\redStateDim$-th eigenvalue is truly greater than its successor.
Instead of solving an eigenvalue problem of the covariance matrix~$\covariance$, a singular value decomposition of the snapshot matrix~$\snaps$ yields a solution to~\eqref{eq: variance opt problem} as well.

The first~$\redStateDim$ (most important) basis vectors are assembled in the projection matrix
\begin{align*}
	\redMatrix_{\text{PCA}} = \begin{bmatrix}
		\redBasis_1, & ... & \redBasis_\redStateDim
	\end{bmatrix}
\end{align*}
so that the reduction mapping results in a projection~$\reduction_{\text{PCA}}(\state)=\redMatrix_{\text{PCA}}^T\state$ and the reconstruction mapping in the corresponding back projection~$\reconstruction_{\text{PCA}}(\redState)=\redMatrix_{\text{PCA}}\redState$.

\subsubsection{Kernel Principal Component Analysis}

Kernel principal component analysis was introduced in \cite{SchoelkopfSmolaMueller97} as a nonlinear extension of \PCA. A more recent overview can be found in~\cite{GarciaGonzalez2020}. The underlying idea is to transform the state onto a higher dimensional feature space~$\featureSpace\subseteq\Rdim^\featureDim$ where it is more likely to obtain linear separability.  
% before performing a \PCA.

The transformation is done using an arbitrary nonlinear map
\begin{align*}
	\featureMap: \state \to \featureMap(\state) \in \featureSpace
\end{align*}
with a very large dimension~$\featureDim \gg \stateDim$. Consequently, the transformed solution manifold~$\solutionManifold_{\featureSpace}=\featureMap(\solutionManifold_{\paramSpace})=\{\featureMap(\state(\param)) \ \vert \ \param\in\paramSpace\}\subseteq\Rdim^\featureDim$ should be better approximable by a linear combination of subspaces than the original space was. 
All that remains to be accomplished is to perform the \PCA on the transformed snapshot matrix
\begin{align*}
	\snaps_{\featureSpace}=
	\begin{bmatrix}
		\featureMap{(\state_{1})}\
		\featureMap{(\state_{2})}\
		...\
		\featureMap{(\state_{\nSnaps})}\
	\end{bmatrix}\in\Rdim^{\featureDim \times \nSnaps}.
\end{align*} 

However, the computational costs of working in the high-dimensional feature space $\featureSpace$ can be avoided by using the so-called \emph{kernel trick}. It enables to work in the original rather than in the feature space by replacing scaler products~$\featureMap(\state_i)^T\featureMap(\state_j)$ with a suitable kernel function 
\begin{align*}
		\kernel(\state_i, \state_j)=(\featureMap(\state_i)\cdot\featureMap(\state_j))=\featureMap(\state_i)^T\featureMap(\state_j).
\end{align*}
As it is possible to formulate the \PCA in such a way that all vectors $\featureMap(\state)$ only appear within scalar products no computations in the feature space have to be solved.
 Applying the kernel function to all samples yields the kernel matrix $\kernelMatrix$ which entries are defined as
\begin{align*}
	\kernelMatrix_{i,j}=\kernel(\state_i, \state_j).
\end{align*}
The kernel matrix must usually be centered as a selected kernel function~$\kernel(\cdot, \cdot)$ does, in general, not result in a centered matrix. 

In contrast to \PCA, the principal components themselves
\begin{align*}
	\redBasis_\redIterator=\sum_{i=1}^{\nSnaps} {\kpcaCoef}_{i\redIterator}\featureMap(\state_i), \quad \redIterator=1,...,\redStateDim
\end{align*}
are not calculated in \KPCA but only the projections of the states onto those components
\begin{align*}
	\redBasis_\redIterator^T\featureMap(\state)=\left(\sum_{i=1}^{\nSnaps} {\kpcaCoef}_{i\redIterator}\featureMap(\state_i)\right)^T\featureMap(\state),
\end{align*}
where~$\featureMap(\state_i)^T\featureMap(\state)$ simply corresponds to the elements~$\kernelMatrix_{i}$ of the kernel $\kernelMatrix$.
The coefficients~${\kpcaCoef}_{i\redIterator}$ can be obtained by solving an eigenvector equation of the kernel matrix~$\kernelMatrix$. 
With the prescribed approach, the reduction mapping of the \KPCA yields 
\begin{equation*}
	\reduction_{\text{kPCA}}(\state)
	% =\redMatrix_{\text{kPCA}}^T\featureMap(\state)=
	=\sum_{\redIterator=1}^{\redStateDim}\redBasis_\redIterator^T\featureMap(\state).
	% =\sum_{\redIterator=1}^{\redStateDim}
\end{equation*}
The reconstruction mapping~$\reconstruction_{\text{kPCA}}(\redState)$ is usually obtained via kernel ridge regression~\cite{BakirWestonSchoelkopf04} as the principal components themselves are never calculated explicitly.
%%%%%%%%%%%%%%%%%%%%%%%%%%%%%%%%%%%%%%%%%%%%%%%%%%%%%%%%%%%%%%%%%%%%%%%%%%%%%%%%%%%%%%%%%%
\subsubsection{Autoencoder}
An Autoencoder (\AE) is a special type of artificial neural network (\NN), which aims to learn the identity function with the constraint that it possesses a bottleneck, see e.g., \cite{LeeCarlberg2020}.
Neural networks in general try to approximate a relationship between an input variable~$\mlInputs$ and the corresponding output variable~$\mlOutputs$ based on an available dataset~$\dataset=\{ (\mlInputs_n, \mlOutputs_n) \}_{n=1}^{\nSnaps}$.
%\begin{figure}
%	\centering
%	\includegraphics[width=\linewidth]{fig/methods/neuron/neuron_tex.pdf}
%	\caption{An artificial neuron of a neural network. Each input~$\mlInput_i^{(l-1)}$ of the~$j$-th neuron of the~$l$-th layer is weighted by a corresponding weight~$\weight^{(l)}_{ji}$ and linearly combined along with the bias~$\weight^{(l)}_{j0}$. This sum is passed through a nonlinear activation function, which serves as output of the neuron}
%	\label{fig:neuron}
%\end{figure}

In the simplest case, their architecture is based on successive layers with neurons. Each neuron is represented as a nonlinear transformation of the linear combination of its inputs. For the~$\neuronIterator$-th  neuron of the~$\layerIterator$-th layer, this can be written as
\begin{equation*}
	\label{eq: NN basis function}
	\mlInput_\neuronIterator^{(\layerIterator)}(\mlInputs^{(\layerIterator-1)})=\activationFunction^{(\layerIterator)}(\sum_{i=1}^{\nUnits^{(\layerIterator-1)}}\weight_{i\neuronIterator}^{(\layerIterator)}\mlInput_i^{(\layerIterator-1)}+\weight_{0\neuronIterator}^{(\layerIterator)})
\end{equation*}
where~$\nUnits^{(\layerIterator-1)}$ describes the number of units in the preceding layer and~$\weight_{ij}, \ i=0,...,\nUnits^{(\layerIterator-1)}$ represent the adjustable weights with the bias~$\weight_{0j}$. Moreover,~$\activationFunction(\cdot)$ is a (nonlinear) activation function. 
Assembling multiple neurons in one layer, the output of the~$\layerIterator$-th layer itself results in  
\begin{equation*}
	\mlInputs^{(\layerIterator)}=\bm{\activationFunction}^{(\layerIterator)}(\Weights_{\ne 0}^{(\layerIterator)}\mlInputs^{(\layerIterator-1)} + \weights_0^{(\layerIterator)}),
\end{equation*}
where the nonlinear activation function~$\bm{\activationFunction}(\cdot)$ is applied element-wise and the adjustable weights are ~$ \Weights^{(\layerIterator)}= \{\Weights_{\ne 0}^{(\layerIterator)}, \weights_0^{(\layerIterator)}\}$.
The networks output is represented by the output of the last layer, i.e., $\mlOutputs=\mlInputs^{(\nLayers)}$.

A basic neural network $\network$ can accordingly be seen as a series of successive transformations\\
\begin{align}
	\network:\ &(\mlInputs, \Weights) \to 
	\label{eq:basic nn}\\
	& \bm{\activationFunction}^{(\nLayers)}(\cdot, \Weights^{(\nLayers)}) \circ \hdots \circ 
	\bm{\activationFunction}^{(1)}(\mlInputs, \Weights^{(1)}) \nonumber
\end{align}
of linear combinations of input variables.

During training, the estimated outputs~$\tilde{\mlOutputs}$ are predicted in the forward pass with fixed parameters~$\Weights$, whereupon the errors are propagated back to adjust the weights with respect to a certain loss function like the mean squared error
\begin{equation} 
	\loss=\frac{1}{\nSnaps}\sum_{i=1}^{\nSnaps}\color{black}(\mlOutputs_i-\tilde{\mlOutputs}_i)^2.
	\label{eq:loss function}
\end{equation} 
In case of gradient descent optimization, the weights are updated
%\begin{equation}
%	%\beta_{kl}^{r+1}=\beta_{kl}^r-\gamma_r\sum_{i=1}^N\frac{\partial L}{\partial \beta_{km}^r}
%	\weights^{r+1}=\weights^r-\learningRate\nabla\loss(\weights^r).
%\end{equation}
in direction of their negative gradient, see \cite{HastieTibshiraniFriedman17}.
%\JKcomment{this paragraph is very basic and can probably removed to shorten the paper if necessary}
% 
\paragraph*{Fully-connected Autoencoder}
% nonlinear ROM: \cite{FrescaDedeManzoni2020}\\
% convolutional recurrent autoencoder \cite{GonzalezBalajewicz2018}\\
% convolutional autoencoder for nonlinear manifolds \cite{LeeCarlberg2020}\\
% graph convolutional autoencoder with mesh simplification \cite{RanjanEtAl2018}\\
% autoencoder coupled with pca \cite{ShenEtAl2021}
 
\begin{figure}
	\begin{subfigure}[t]{216.24094pt}
		\centering
		\includegraphics[]{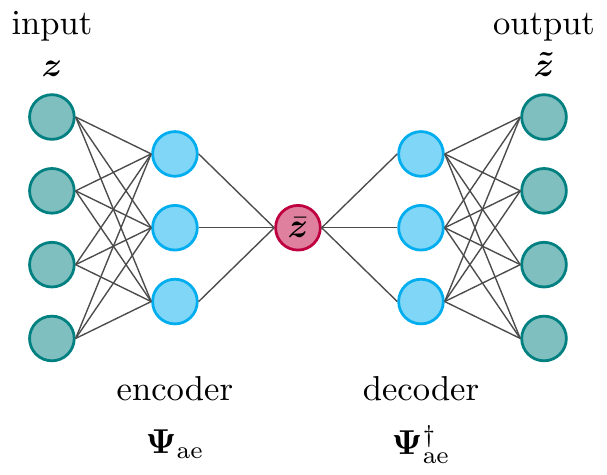}
		\caption{classical autoencoder}
		\label{fig: classical autoencoder}
	\end{subfigure}
	\begin{subfigure}[t]{216.24094pt}
		\includegraphics[]{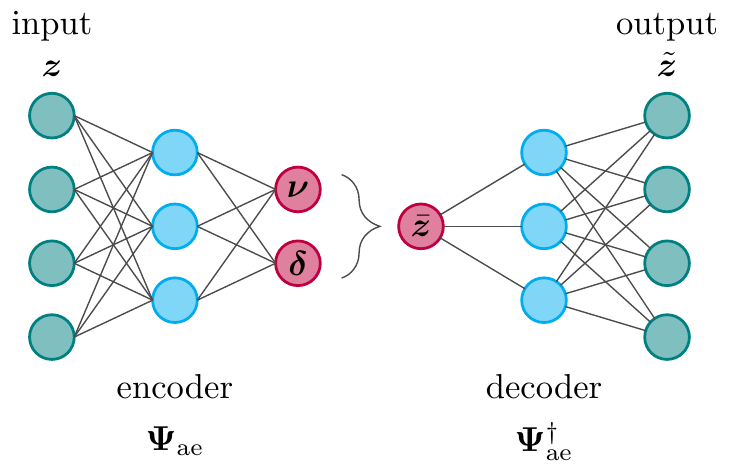}
		\caption{variational autoencoder}
		\label{fig: variational autoencoder}
	\end{subfigure}
	\caption{Schematic representation of autoencoder architectures.}
	\label{fig:autoencoder}
\end{figure}

Autoencoders have already been introduced as a nonlinear alternative to \PCA in the early 90's \cite{Kramer1991}.
As previously mentioned, they aim to learn the identity mapping~$\network: \stateSpace \to \stateSpace$ with~$\stateApprox=\network(\state)$ but with a bottleneck in their layer size as exemplified in Fig.~\ref{fig: classical autoencoder}. This architecture ensures that the network learns a reduced representation of the input variables within the network. Consequently, the autoencoder's input~$\mlInputs=\state$ as well as the output~$\mlOutputs=\state$ correspond to the system state for the considered problem.

An \AE consists of two subnetworks: the encoder~$\reductionAE$ and the decoder~$\reconstructionAE$. In the encoder~$\reductionAE: \stateSpace \to \redStateSpace$, the state dimension is decreased from the original system size $\stateDim$ to $\redStateDim$, i.e.,~$\redState=\reductionAE(\state)$. Analogous to equation~\eqref{eq:basic nn}, a encoder with $\nLayers$ layers can be represented by the function composition
\begin{align*}
	\reductionAE: \ &(\state; \Weights_\text{enc}) \to\\ &
	\bm{\activationFunction}_\text{enc}^{(\nLayers)}(\cdot, \Weights_\text{enc}^{(\nLayers)}) \circ \hdots \circ 
	\bm{\activationFunction}_\text{enc}^{(1)}(\state, \Weights_\text{enc}^{(1)}).
\end{align*}
The decoder in turn is a mapping from the reduced to the original space ~$\reconstructionAE: \redStateSpace \to \stateSpace$ with~$\stateRec = \reconstructionAE(\redState)$ which again is a series of successive transformations
\begin{align*}
	\reconstructionAE: \ &(\redState; \Weights_\text{dec}) \to \\ & 
	\bm{\activationFunction}_\text{dec}^{(\hat{\nLayers})}(\cdot, \Weights_\text{dec}^{(\hat{\nLayers})}) \circ \hdots \circ 
	\bm{\activationFunction}_\text{dec}^{(1)}(\redState, \Weights_\text{dec}^{(1)}) .
\end{align*}

Consequently, the autoencoder results out of the composition of the encoder and decoder network
\begin{align*}
	\autoencoder: \state \to \reconstructionAE \circ \reductionAE(\state).
\end{align*}
The optimal trainable network weights~$\Weights_{\text{ae}}^* = \{\Weights_\text{enc}^*, \Weights_\text{dec}^*\}$ with respect to the loss function~\eqref{eq:loss function} result from 
\begin{align*}
	\Weights_{\text{ae}}^* = \arg \min_{\Weights_{\text{ae}}} \loss(\autoencoder(\state; \Weights_{\text{ae}}), \state).
\end{align*}

One problem in classical autoencoders is that they can lack regularization of the reduced/latent space. Hence, points that are close to each other in the reduced space can result in very different reconstructions. Moreover, the reduced space is not easy to interpret and cannot be assigned to any concrete effects in the full space. To circumvent these problems variational autoencoders~\cite{Kingma2013, Rezende2014} can be used.

\paragraph*{Variational Autoencoder}

The input of variational autoencoders~(\VAEs) is in contrast to the regular autoencoders not transformed into a single reduced coordinate but into a distribution over the reduced space. Often, the prior probability distribution (\emph{prior}) over the reduced/latent variables $\vaePrior(\redState)$ is assumed to be Gaussian~$\mathcal{N}(\bm{0}, \bm{I})$ with zero mean and unit variance. 
Hence, the encoder network returns a mean value~$\vaeMean(\state)$ and a variance~$\vaeVar(\state)$ describing a normal distribution~$\mathcal{N}(\vaeMean, \vaeVar)$, see~\figref{fig: variational autoencoder}. 
Correspondingly, its output $\begin{bmatrix}\vaeMean^T \ \vaeVar^T\end{bmatrix}^T\in\Rdim^{2\redStateDim}$ is twice the size of the reduced coordinates~$\redStateDim$.

To reconstruct the original data, a sample is drawn from the reduced/latent distribution~$\redState \sim \mathcal{N}(\vaeMean, \vaeVar)$ and then transformed by the decoder network. 
As a result, the variables in the reduced space have continuity, i.e., close points in the reduced space also give similar reconstructions, and each point within the distribution should give a meaningful reconstruction. 
Consequently, \VAEs are often used in generative tasks where data samples are created based on random samples from the reduced/latent distribution. Nevertheless, they also provide a great tool for dimensionality reduction as demonstrated in~\cite{Mahmud2020}.

In detail, \VAEs deal with the problem of variational inference, i.e., they aim to infer the posterior probability distribution (\emph{posterior})~$\vaePrior(\redState \vert \state)$ over a hidden (in our case reduced/latent) variable~$\redState$ given representative data~$\state$.
This is achieved by maximizing the Evidence Lower-Bound (ELBO)
\begin{align}
	\loss_{\text{\VAE}}=\mathbb{E}_{\vaePosterior(\redState \vert \state)}[\log \vaePrior(\state \vert \redState)] - \vaeDis \text{KL}(\vaePosterior(\redState \vert\state)\vert\vert  \vaePrior(\redState))
	\label{eq: elbo}
\end{align}
where the first part serves as reconstruction loss. It ensures that the log-probability of~$\state$ given~$\redState$ drawn from the approximated posterior $\vaePosterior(\redState \vert \state)$ is maximized, i.e., that the input is reconstructed well from a sample in the reduced space. 
The second part includes the Kullback-Leibler divergence~$\text{KL}$, which is a distance measure defined on probability distributions. This ensures that the reduced coordinates respect the assumed prior over the latent variables $\vaePrior(\redState)$, i.e, that the approximated posterior $\vaePosterior(\redState \vert \state)$ is pushed closer to the prior. 
With the parameter~$\vaeDis$, the $\text{KL}$ divergence can be weighted depending on the requirements to either focus more or less on the reconstruction ability of the \VAE. A large value of~$\vaeDis$ leads to the so-called disentanglement of the latent variables so that a single unit of the variable reacts sensitive to changes in individual system characteristics. Thereby, the impact of this unit of the variable on the reconstructions can be better interpreted. 

In \VAEs, the variational inference problem is solved by parameterizing the likelihood and posterior as functions by neural networks. 
Specifically, the likelihood function is a function parameterized by the decoder $\stateApprox=\reconstructionAE(\redState) \sim \vaePrior(\state\vert\redState)$ reconstructing the full state~$\state$ given a reduced state~$\redState$ and the approximated posterior~$\vaePosterior(\redState \vert \state)$ is represented by the encoder network~$\redState=\reductionAE(\state) \sim \vaePosterior(\redState \vert \state)$. During training of the model, the so-called reparameterization trick is used, to enable backpropagation of the resulting loss through the network since the process of sampling from a distribution defined by the encoder~$\reductionAE$ would normally not be differentiable, see~\cite{Kingma2013, Rezende2014}.
Hence, a noise term~$\vaeNoise\sim\mathcal{N}(\bm{0},\bm{I})$, which is not parameterized by the network, is introduced whereby a sample can be drawn following
\begin{align*}
	\redState=\vaeMean(\state) + \vaeNoise \vaeVar(\state).
\end{align*}
By doing so, the ELBO function~\eqref{eq: elbo} can be optimized using gradient descent.
During evaluation of the model, we neglect the predicted variance~$\vaeVar$ and only use the predicted mean~$\vaeMean$ as output of the encoder.

% `In \VAEs a model of the form~$\vaePrior(\state, \redState)=\vaePrior(\redState)\vaePrior(\state \vert \redState)$ is trained, where~$\vaePrior$ represents a prior distribution over the reduced / latent states~$\redState$ and the encoder represents the likelihood function  $\stateApprox=\reconstructionAE(\redState) \sim \vaePrior(\state\vert\redState)$ that reconstructs~$\state$ given a reduced state~$\redState$. As the posterior~$\vaePrior(\redState \vert \state)$ usually is intractable, a posterior approximation~$\vaePosterior(\redState \vert \state)$ is used instead and can be interpreted as probabilistic encoder $\redState=\reductionAE(\state) \sim \vaePosterior(\redState \vert \state)$.
One major problem that arises within the context of fully-connected \AEs is the very high number of trainable parameters, i.e., weights $\Weights$, compared to the already presented methods. This leads to an expensive training phase and is usually mitigated by using \emph{convolutional autoencoders} as in \cite{LeeCarlberg2020, GonzalezBalajewicz2018}. Convolutional neural networks (\CNNs) reduce the number of required model parameters by \emph{parameter sharing} and generalize well by detecting local patterns in spatially distributed data. 
Nevertheless, they are of limited use for the reduction of finite element models as they are especially suited for data that is discretized in a grid-like structure. \FE models, on the contrary, usually possess an irregular spatial discretization and the generalization of \CNNs to such data is not trivial.
Therefore, in this work, we exclusively use fully-connected autoencoders which are still admissible given the system size of the human upper-arm model.
It should be noted, however, that one possible solution to apply \CNNs to \FE models may be found in graph convolutions \cite{DefferrardBressonVandergheynst2016}.

\subsection{Regression}
All presented dimensionality reduction techniques, \PCA, \KPCA, \AE, and \VAE, offer the possibility to identify suitable transformations to represent the system states $\state$ in a low-dimensional subspace and thus form the foundation for further procedure. The reduction not only retains just the most important system features but also enables highly efficient training of regression algorithms. Thereby, the search for a function~$\regression: \paramSpace \to \redStateSpace$  approximating the relationship from simulation parameters to reduced system behavior $\param \to \redState$ is simplified.

As we only rely on data in this work, any regression algorithm can be used for this task. However, data generation is often an expensive process, especially in the case of complex \FE models. Hence, whenever the offline phase is resource-limited, an algorithm that can cope with less data is preferable. 
Since the dataset on which this work is based also only contains relatively few samples, Gaussian process regression is the algorithm of choice. It is well-known to perform well, even in data-poor regimes and was successfully applied in other publication~\cite{KneiflHayFehr2021, KapsCzechDuddeck2022} in the context of \LORE.

\subsubsection{Gaussian Process}
% stick to Advanced_... Rassmussen
The basic idea of \emph{Gaussian Processes} (\GPs) is to limit the space of functions fitting the given data by imposing a prior on the functions first and condition them with available observations to receive the posterior probability distribution, see \cite{Rasmussen03}.  
This approach relies on multivariate Gaussian distributions which are defined by a mean~$\vaeMean$ and a covariance matrix~$\covMatrix$. According to this, Gaussian Processes are completely defined by a mean function~$\meanFunc(\mlInputs)$ and a covariance or rather kernel function~$\kernelFunc(\mlInputs_i,\mlInputs_j)$ for some inputs $\mlInputs$. Thus, a \GP can be defined as
% Gaussian process
\begin{equation*}
	\bm{f} \sim \mathcal{N}(\meanFunc, \kernelMatrix),
\end{equation*}
where~$\kernelMatrix$ is the kernel matrix obtained by applying the kernel function element-wise to all inputs. 
The selected kernel function implies a distribution over functions, giving a higher probability to those satisfying certain properties such as smoothness to a greater extent. In order to update the prior with training data, let~$\mlInputs$ be training points with an observed result~$\mlOutputs$ and a mean function~$\meanFunc_1$ and~$\mlInputs_*$ test points with the mean function~$\meanFunc_2$ for which the output~$\mlOutputs_*$ shall be predicted. Suppose~$\mlOutputs$ and~$\mlOutputs_*$ are jointly Gaussian  
% joint distribution
\begin{equation*}
	\begin{bmatrix} \mlOutputs \\ \mlOutputs_*\end{bmatrix} \sim 
	\mathcal{N}\begin{pmatrix}
		\begin{bmatrix}\meanFunc_1 \\\meanFunc_2\end{bmatrix},
		\begin{bmatrix}\kernelMatrix & \kernelMatrix_*\\ \kernelMatrix_*^T & \kernelMatrix_{**}\end{bmatrix}
	\end{pmatrix},
\end{equation*}
where~$\kernelMatrix$ is the kernel matrix for the variables in training input space,~$\kernelMatrix_*$ between the training and test data, and~$\kernelMatrix_{**}$ among the test variables themselves. Then, the posterior can be calculated following
% posterior 
\begin{align*}
	% \label{eq: GP}
%	\bm{f}_{* | \bm{x}_{*}, \bm{x}, \bm{f}}
	\mlOutputs_{*} \vert \mlInputs_{*}, \mlInputs, \bm{f} \sim 
	\mathcal{N}(\meanFunc_p ,\ \kernelMatrix_p)
\end{align*} 
with mean~$\meanFunc_p={\meanFunc_2+\kernelMatrix_{*}^T\kernelMatrix^{-1}(\mlOutputs-\meanFunc_1)}$ and~$\kernelMatrix_p={\kernelMatrix_{**}-\kernelMatrix_{*}^T\kernelMatrix^{-1}\kernelMatrix_{*}}$.
For regression, the \GP's mean function~$\meanFunc_p$ serves as predicted output. 
During training, the kernel function's hyperparameters are tuned to obtain the final regression algorithm~$\regression$.
For further information on \GPs please rely on \cite{Rasmussen03}. 
%Hence, it is possible to describe the system dynamics in their reduced representation~$\redState(t)$ as long as the reconstruction of the system states~$\reconstructionLin\reductionLin\state(t)$ satisfies the approximation requirements.
%
\section{Implementation}
\label{sec:implementation}
%\JKcomment{not happy with the section title}
The presented framework consists of two successive steps. The first of which is a coordinate transformation to obtain a low-dimensional representation from high-dimensional data and the second is the approximation of the system behavior in the reduced space.
For the corresponding tasks, individual algorithms have been presented. However, the entire framework is structured generically so that a wide variety of further suitable algorithms exists. Therefore, in the following the variables~$\reduction$ and~$\reconstruction$ are used for reduction and reconstruction in general.

For the implementation of \PCA, \KPCA, and \GP it is relied on the software library scikit-learn~\cite{scikit}, whereas the autoencoders are implemented using TensorFlow \cite{tensorflow2015}. 
The high-fidelity simulations were conducted using the commercial FEM tool LS-DYNA and the resulting data is provided to the interested reader in~\cite{darusArm}. This dataset contains the model's muscle activations, displacements, as well as processed von Mises stress. The toolbox to generate the presented surrogate models, however, is still in progress and will be published at a later date. 

%%%%%%%%%%%%%%%%%%%%%%%%%%%%%%%%%%%%%%%%%%%%%%%%%%%%%%%%%%%%%%%%%%
%% 							ALGORITHM  							%%
%%%%%%%%%%%%%%%%%%%%%%%%%%%%%%%%%%%%%%%%%%%%%%%%%%%%%%%%%%%%%%%%%%
\subsection{Algorithm}
The algorithm can be divided into an offline and an online phase. While data creation and surrogate modeling take place in the former one, the evaluation of the surrogate for new unseen parameters is part of the latter one. 

In the \emph{offline phase}, which is summarized in Algorithm~\ref{alg: Offline}, a finite set of simulation parameter~$\paramSpace_\nSnaps=\{\param_1,...,\param_{\nSnaps}\}$ from the parameter domain~$\paramSpace$ is created using some sampling strategy. Based on the selected parameter, the high-fidelity model is evaluated and its results are assembled for further processing.
The following steps of the offline phase involve (i) calculation of coordinate transformation for dimensionality reduction and subsequently (ii) fitting of the regression algorithm to approximate the reduced system behavior.

To solve the task of dimensionality reduction, the high-fidelity simulation results are assembled in the dataset
\begin{align}
	\dataset_{\text{red}}=\{(\state_n, \state_n) \}_{n=1}^{\nSnaps}
	\label{eq: reduction dataset},
\end{align} which contains the full states as in- and output and can be rewritten as snapshot matrix~\eqref{eq: snapshot matrix}
Based on this data, the reduction mapping~$\reduction$ and the reconstruction mapping~$\reconstruction$ are fitted. Once these mappings are found the regression dataset can be composed by reducing the system states~$\redState=\reduction(\state)$ resulting in
\begin{align}
	\dataset_{\text{regr}}=\{ (\param_n, \redState_n) \}_{n=1}^{\nSnaps}
	\label{eq: regression dataset}.
\end{align} 
It is composed of the individual simulation parameters, i.e., the muscle activations as inputs and the corresponding reduced system states obtained by the reduction algorithms as output. Subsequently, the regression algorithm $\regression$ is trained using $\dataset_{\text{regr}}$ as data basis.
The composition of the fitted reconstruction mapping $\reconstruction$ along with the regression model~$\regression$ yields the complete surrogate model 
\begin{align*}
	\stateApprox = \reconstruction \circ \regression(\param).
\end{align*}
It is used in the \emph{online phase} to predict the system behavior, see Algorithm~\ref{alg: Online}.
In other words, the regression model is evaluated on new (unseen) simulation parameters to receive an approximation of the respective reduced states followed by a transformation back into the high-dimensional physical space.
For an overview of the entire workflow including the offline as well as the online phase of the algorithm please refer once again to \figref{fig:low-rank regression}.
\begin{algorithm}
	% \caption{\textsf{\LRR Offline }}
	\caption{}
	\textbf{Input}: parameter domain~$\paramSpace$, amount of simulations~$\nSnaps$\\
	\textbf{Output}: reduction algorithm~$\reduction$, regression model~${\regression}$
	\begin{algorithmic}[1]
		\State create parameter set~$\paramSpace_\nSnaps=\{\param_1,...,\param_{\nSnaps}\}$
		\For {$j=1,...,\nSnaps$}
		\State evaluate high-fidelity model~$\Flow(\param_j)$
		\EndFor
		\State assemble dataset~$\dataset_{\text{red}}$
		\State calculate reduction transformation~$\reduction(\state)$ via \PCA/\KPCA/\AE 
		\State assemble dataset~$\dataset_{\text{regr}}$
		\State train regression algorithm~${\regression}(\param; \weights)$
	\end{algorithmic}
	\label{alg: Offline}
\end{algorithm}
\begin{algorithm}
	\caption{}
	% \caption{\textsf{\LRR Online}}
	\textbf{Input}:  parameter~$\param\in\paramSpace$\\
	\textbf{Output}: approximated state~$\stateApprox$ \nonumber
	\begin{algorithmic}[1]
		\State predict reduced state~$\redStateApprox=\regression(\param)$
		\State transform into full space~$\stateApprox=\reconstruction(\redStateApprox)$
		\State \textbf{return}~$\stateApprox$
	\end{algorithmic}
	\label{alg: Online}
\end{algorithm}
\subsection{Error Quantities}
It is crucial to not only consult the complete surrogate model but also its components to validate its overall approximation quality and gain insights into the individual sources of error. Hence, besides the overall approximation error, also the error induced by the reduction and the regression algorithm are investigated. 

For this purpose, three relative performance scores are introduced which allow a reliable assessment. 
% independent of the considered quantity such as deformation or stress. 
In addition, absolute error measurements serve as physical interpretable quantities.
\paragraph*{Relative Performance Scores}
In this paper, different reduction methods and their impact on the approximation are explored. 
A well-suited quantity to consider their impact isolated from the other sources of error is introduced as \emph{reconstruction score} following the calculation rule
\begin{align}
	\sRec(\state, \stateRec)=
	1-\frac{\Vert \state - \stateRec \Vert_2}{\Vert  \state \Vert_2}.
	\label{eq:errorrec}
\end{align} 
It compares a state $\state$ with its reconstruction~$\stateRec=\reconstruction \circ \reduction(\state)$.
This score, as well as the two following ones, reaches its optimum with a value of 1, whereas a value of 0 corresponds to no prediction at all. It is to be noted that~$\sRec$ can become negative as the reconstruction can be arbitrarily poor.

Similar to the isolated reconstruction score, the \emph{regression score}
\begin{align}
	\sRegr(\redState, \redStateApprox)=
	1-\frac{\Vert \redState - \redStateApprox \Vert_2}{\Vert  \redState \Vert_2}
	\label{eq:errorred}
\end{align} 
validates the performance only of the regression algorithm by comparing the prediction~$\redStateApprox=\regression(\param)$ with the reduced reference state~$\redState$. This is the only performance measurement taking place in the reduced space while all other measurements are calculated in the full physical space.

The composition of both parts, i.e. the overall approximation quality, is measured by the so-called \emph{approximation score}
\begin{align}
	\sApprox(\state, \stateApprox)=
	1-\frac{\Vert \state - \stateApprox \Vert_2}{\Vert  \state \Vert_2},
	\label{eq:errorapprox}
\end{align}
where~$\stateApprox=\reconstruction(\regression(\param))$ is the overall approximation of the reference~$\state$.
To visualize the individual scores, the connection between the compared quantities is shown in Fig.~\ref{fig:low-rank regression}.

While the presented scores can all be assessed for a single simulation, their mean value over all simulations
\begin{align*}
	\hat{s}_\text{rec/regr/approx}(\cdot,\cdot)=\frac{1}{\nSnaps}\sum_{i=1}^{\nSnaps}s_\text{rec/regr/approx}(\cdot,\cdot)
%	\label{eq:mean score}
\end{align*}
is used to compare several simulations at once. %\JKcomment{any ideas how to write down this equation more formally?}

\paragraph*{Absolute Error Measures}
In addition, to the relative scores, absolute error measures are consulted for the overall approximation. Therefore, the sample wise 2-norm 
\begin{align}
	\eTwo(\state_\nodeIterator, \stateApprox_\nodeIterator) = \lVert \state_\nodeIterator(\param)-{\stateApprox}_\nodeIterator(\param) \rVert_2 
\end{align} 
of the error between the features of the $\nodeIterator$-th node/element~$\state_\nodeIterator$ and their approximation~$\stateApprox_\nodeIterator$ is used. 
Furthermore, its mean 
\begin{align}
	\eTwoMean = \frac{1}{\nItem}\sum_{\nodeIterator=1}^{\nItem} \eTwo(\state_\nodeIterator, \stateApprox_\nodeIterator)
	\label{eq: e2 mean}
\end{align} 
and maximum value
\begin{align}
	\eTwoMax = \max_{m\in\{1,...,\nItem\}} \eTwo(\state_\nodeIterator, \stateApprox_\nodeIterator)
	\label{eq: e2 max}
\end{align} 
over all nodes/elements are used for comparisons.
Note that in case of displacements, where $\state_\nodeIterator=\disps_\nodeIterator$ corresponds to the coordinates of a node, the 2-norm~$\eDist(\disps_\nodeIterator, \tilde{\disps}_\nodeIterator)=\eTwo(\disps_\nodeIterator, \tilde{\disps}_\nodeIterator)$ represents the Euclidean node distance of the~$\nodeIterator$-th node~$\disps_\nodeIterator$ and its approximation~$\tilde{\disps}_\nodeIterator$. \\
%  $\nItem=\nNodes$\\
For the element-wise scalar valued stress samples, $\eStress(\stressScalar_\nodeIterator, \tilde{\stressScalar}_\nodeIterator)=\eTwo(\stressScalar_\nodeIterator, \tilde{\stressScalar}_\nodeIterator)$ just matches the absolute difference between the von Mises stress of the~$\nodeIterator$-th element and its approximation~$\tilde{\stressScalar}_\nodeIterator$.
%  for all elements, i.e.~$\nodeIterator=1,...,\nItem=\nNodes$.\\
\\
As last performance measure, the required computational time for one sample~$\compTime$ is used to account for the resource savings achieved by the surrogate models.
% As the last performance measure the simulation speedup
% \begin{align*}
% 	\speedup= \frac{\compTime_{\Flow(\cdot)}} {\compTime_{\regression(\cdot)}}.
% \end{align*} 
% is used to account for the acceleration of computational time achieved by the surrogate models.
%\JKcomment{will we use only computational time or also the speedup as we do not have numbers nor the correct hardware?}

\subsection{Surrogates for Displacement and Stress}
The implementation of a surrogate model that captures the displacements of the model and one that captures the stress is very similar and follows the presented approach in both cases. Nevertheless, both quantities, i.e., $\disps$ and $\stress$ live in such different spaces that each requires its own reduction algorithm. Hence, for each quantity an individual dataset must be assembled following~\eqref{eq: reduction dataset}. This yields $\dataset_{\text{red}}^{\text{disp}}=\{(\disps_n, \disps_n) \}_{n=1}^{\nSnaps}$ for the displacements and $\dataset_{\text{red}}^{\text{stress}}=\{(\stress_n, \stress_n) \}_{n=1}^{\nSnaps}$ for the von Mises stress.\\
For the subsequent regression, a single algorithm could be used to predict values of both quantities. However, we use a single regression algorithm per variable for the sake of clarity. This, in turn, requires individual data sets $\dataset_{\text{regr}}^{\text{disp}}=\{ (\param_n, \bar{\disps}_n) \}_{n=1}^{\nSnaps}$ and $\dataset_{\text{regr}}^{\text{stress}}=\{ (\param_n, \bar{\stress}_n) \}_{n=1}^{\nSnaps}$ for the approximation of the reduced displacements~$\bar{\disps}$ and reduced stress~$\bar{\stress}$.

\section{Results}\label{sec:results}

%\begin{figure*}
%	\includegraphics[width=\textwidth,height=4cm]{fig/results/coordinates/node_displacements_small_v2.png}
%	\caption{This is a tiger.}
%\end{figure*}

% This is a test 
For the creation of the high-fidelity, i.e., finite element simulation results, 1053 muscle activations~$\param\in\Rdim^{\muscleDofs}$ are equidistantly sampled on the parameter domain~$\paramSpace=[0,1]\times[0,1]\times[0,1]\times[0,1]\times[0,1]$ as simulation parameters. A value of~1 corresponds to full muscle activation, whereas a value of~0 corresponds to no muscle activation. The amount of samples was chosen to be rather small to reduce the overall number of FE model evaluations, since calculating one activation state can take minutes on a high-performance computing system with 64 processors in parallel. %\JKcomment{Repitition. Remove this sentence? Is this correct? How were they scaled?}
Based on these muscle activations, the finite element simulation was evaluated, resulting in 1053 samples for displacement~$\disps$ and stress~$\stress$, which are used to build the training data sets following~\eqref{eq: reduction dataset} to fit the models. Another~$\nSamplesTestNum$ parameters~$\param$ are randomly sampled, using a uniform distribution, on the parameter domain. These serve as a test set. %\JKcomment{How was the test data sampled?}.
%Compared to other similar applications and the dimension of the input, a 
A particular challenge in surrogate modeling for this simulation lies in the relatively poor data basis, due to its prohibitive computational cost. 
The following results (with exception of the high-fidelity simulations) were produced on an Apple M1 Max with a 10-Core CPU, 24-Core GPU, and 64 GB of RAM.

%\JKcomment{Should we mention the following paragraph more early? Maybe even in the motivation?}
In our previous studies~\cite{KneiflGrunertFehr21, KneiflHayFehr2021}, linear reduction methods have proven to be suitable for the low-dimensional representation of displacements obtained by and elastic multibody and \FE models. 
A similar conclusion can be reached for the arm model by considering the singular values~$\singvalues$, which represent the eigenvalues of the covariance matrix~$\covariance$ of the centered snapshot matrix~$\hat{\snaps}$. The contribution of a reduced basis vector of the \PCA to the total reconstruction of the original quantity is reflected by the magnitude of the corresponding singular value. That means if only a few singular values contribute significantly, it is possible to represent a quantity with only a few reduced basis vectors without losing much information. 
If this is in contrast not the case, the reduction either requires numerous reduced basis vectors or introduces a remarkable error. 
This is closely related to the Kolmogorov $n$-width~(\cite{UngerGugercin2019}, \cite{LeeCarlberg2020})
\begin{align*}
	d_n(\Flow(\paramSpace))
	% 	:= \underset{\mathcal{Y}_n\leq\mathcal{Y}}{\inf}d(\mathcal{Y}_n, \mathcal{S}(\mathbb{W}))
	:=\underset{\stateSpace_{n} \subseteq \stateSpace}{\inf} \ 
		\underset{\state\in\solutionManifold_{\paramSpace}}{\sup} \ 
			\underset{\stateApprox_n\in\stateSpace_{n}}{\inf}
			\| \state-\stateApprox_n \|
\end{align*}
which quantifies the optimal linear trial subspace by describing the largest distance between any point in the solution manifold for all parameters~$\solutionManifold_{\paramSpace}$ and all $n$-dimensional subspaces~$\stateSpace_{n}\subseteq \stateSpace$.
Supposing that the given system has unique solutions for all parameter samples, the intrinsic dimension of the solution space is at the most equal to the number of parameters, or number of parameters plus one for dynamical systems. 

\begin{figure}[h]
	\includegraphics[]{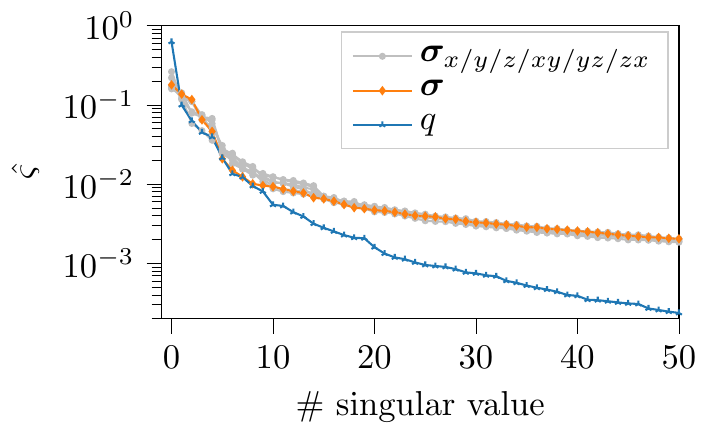}
	\caption{Scaled singular values of displacement and stress data for \PCA as indicator of the Kolmogorov $n$-width.}
	\label{fig: sing values}
\end{figure}

Consequently, if a linear subspace were sufficient to represent a given system, i.e., if the system possessed a fast decaying Kolmogorov $n$-width, we would expect that using as many PCA modes as the intrinsic dimension leads to a satisfying low-dimensional representation of the system. 
For the given human arm model which is parameterized by a five-dimensional parameter space~$\paramSpace\subseteq\Rdim^{5}$, this means that having the most dominant behavior within the first five modes would justify the use of a linear subspace. 
The coarse of the singular values $\hat{\singvalues}$ is considered in \figref{fig: sing values} to investigate the suitability of a linear subspace.
For a suitable presentation, the singular values are scaled by the sum of singular values, i.e. 
$\hat{\singvalue}^{(\redIterator)}={\singvalue^{(\redIterator)}} \mathbin{/} {\sum_{i=1}^{\nSnaps}\singvalue^{(i)}}$.
On the one hand, the first five \PCA modes for the displacements, indeed, capture almost 90\% of the behavior of all data.  
On the other hand, considering the results of the stress data it becomes apparent that the singular values do not decrease as fast as it is the case for the displacements and single singular values are not that dominant. Accordingly, the first five only capture about 60\%. This already indicates that the approximation of the stress is more challenging. 

In this context, it is to be noted that this does not result from considering the von Mises stress instead of single stress components. The individual components~$\stress_{x},\ \stress_{y},\ \stress_{z},\ \stress_{xy},\ \stress_{yz},\ \stress_{zx}$  possess a similar behavior as shown in \figref{fig: sing values}. Accordingly, it seems to be more difficult to represent the stress data in a low-dimensional linear subspace in general compared to the displacements. Furthermore, we expect that our results for the von Mises stress are transferable to the individual components. 
% This confirms together with the better clarity the choice of the von Mises stress as quantity of interest.

\paragraph*{Hyperparameters}
Another noteworthy aspect in this context is that a reduced coordinate~$\bar{z}^{(\redIterator)}$ belonging to a reduced basis vector~$\redBasis_\redIterator$ with a relatively small corresponding singular value~$\singvalue^{(\redIterator)}$ tends to possess less assignable characteristics and relations. Thus, they are often more difficult to approximate by regression algorithms. This in turn results in a preference for a few meaningful coordinates over many less meaningful coordinates for the following investigation, not only because of their dimensionality but also because of their predictability. %\JKcomment{This paragraph only results out of my experiences and not from a provable elaboration. Shall we still keep it?}
The reduced system size~$\redStateDim$ of the surrogate models is set to the same value for \PCA, \KPCA, \AE, and \VAE to guarantee a fair comparison. In case of the displacement, the reduced system size is chosen so that a reconstruction score above~$\sRec>0.99$ is achieved using \PCA resulting in a dimension of~$\redStateDim_{\text{disp}}=10$. For stress, on the contrary, it is set to $\redStateDim_{\text{stress}}=13$ resulting in a reconstruction score above~$\sRec>0.95$ using \PCA.

Based on this preselected parameter, the individual hyperparameters of the different reduction algorithms are tuned using a grid search on the $k$-cross-validated training data set~\eqref{eq: reduction dataset}. 
As for \PCA no further hyperparameter than the reduced system size~$r$ exists, only the hyperparameter for~\KPCA, \AE, and \VAE are listed in Table \ref{tab: kpca hyper parameter} and Table \ref{tab: Autoencoder hyper parameter}, respectively. Both autoencoders share the same underlying architecture to ensure a fair comparison.
%\the\textwidth
\begin{table}
	\setlength{\tabcolsep}{3pt}
	\centering
	\caption{\KPCA algorithms hyperparameter}
	\label{tab: kpca hyper parameter}
	\npdecimalsign{.}
	\nprounddigits{3}
	\begin{tabular}[c]{c c c c c c c c}
		\toprule
		& \makecell{kernel\\function} & $\kernelParam$ & $\kernelBias$ & $\kernelDegree$ & $\kpcaAlpha$\\ 
		\midrule
			disp & polynomial~\eqref{eq: poly kernel} & $1\cdot10^{-10}$ & 452 & 6 & $1\cdot10^{9}$\\ 
			stress & polynomial~\eqref{eq: poly kernel} & $1\cdot10^{-6}$ & 276 & 6 & $1\cdot10^{6}$\\ 
%		\bottomrule
	\end{tabular} 
	\npnoround
\end{table}
\begin{table}
	\setlength{\tabcolsep}{3pt}
	\centering
	\caption{Autoencoder hyperparameter}
	\label{tab: Autoencoder hyper parameter}
	\npdecimalsign{.}
	\nprounddigits{3}
	\begin{tabular}[c]{c c c}
		\toprule
		&\makecell{hidden\\layer} & \makecell{activation\\function~\eqref{eq: selu}} \\ 
		\midrule
			disp & $\reductionAE:75\times50\times40\times30$ & $linear\times selu\times selu\times selu$ \\ 
				& $\reconstructionAE:30\times40\times50\times75$ & $selu\times selu\times selu\times linear$ \\ 
			stress& $\reductionAE:140\times70\times35$ & $selu\times selu\times selu$ \\
			& $\reconstructionAE:35\times70\times140$ & $selu\times selu\times selu$ \\
%		\bottomrule
	\end{tabular} 
	\npnoround
\end{table}

For \KPCA, a polynomial kernel function
\begin{align}
	\kernel_\text{poly}(\state_i, \state_j)=(\kernelParam\state_i^T\state_j+\kernelBias)^\kernelDegree
	\label{eq: poly kernel}
\end{align}
is chosen and the scaled exponential linear unit~(\SELU)
\begin{align}
	\activationFunction_{\text{selu}}(\mlInput)=
	\begin{cases}
		\seluScale\mlInput & \mlInput \geq 0\\
		\seluScale\seluAlpha(e^{\mlInput}-1) & \mlInput < 0\\
	\end{cases}
	\label{eq: selu}
\end{align}
with~$\seluScale=1.051$ and~$\seluAlpha=1.673$ serves as activation function in the autoencoder.

For each fitted reduction algorithm, the dataset~\eqref{eq: regression dataset} is assembled and regression algorithm is trained. 
Usually, the regression task is solved by a subnetwork in the case of autoencoders.
However, we want to focus on the comparison of the individual reduction methods in the following investigation and want to examine them as independently as possible from the regression algorithm. 
Hence, the same regression algorithm is used for \PCA, \KPCA, \AE, and \VAE to enable comparability of their results within the context of \LORE. 
In detail, a Gaussian process regression algorithm with consistent hyperparameters for all three reduction algorithms is used, see Table~\ref{tab: gp hyper parameter}. 
\begin{table}
	\setlength{\tabcolsep}{3pt}
	\centering
	\caption{Gaussian Process regression hyperparameter}
	\label{tab: gp hyper parameter}
	\npdecimalsign{.}
	\nprounddigits{3}
	\begin{tabular}[c]{c c c c}
		\toprule
		\makecell{kernel\\function} & $\kernelParam$ & $\kernelBias$ & $\kernelDegree$\\ 
		\midrule
			 polynomial~\eqref{eq: poly kernel} & $1$ & 1.15& 6\\ 
%		\bottomrule
	\end{tabular} 
	\npnoround
\end{table}

\begin{figure*}
	\centering
	\includegraphics[]{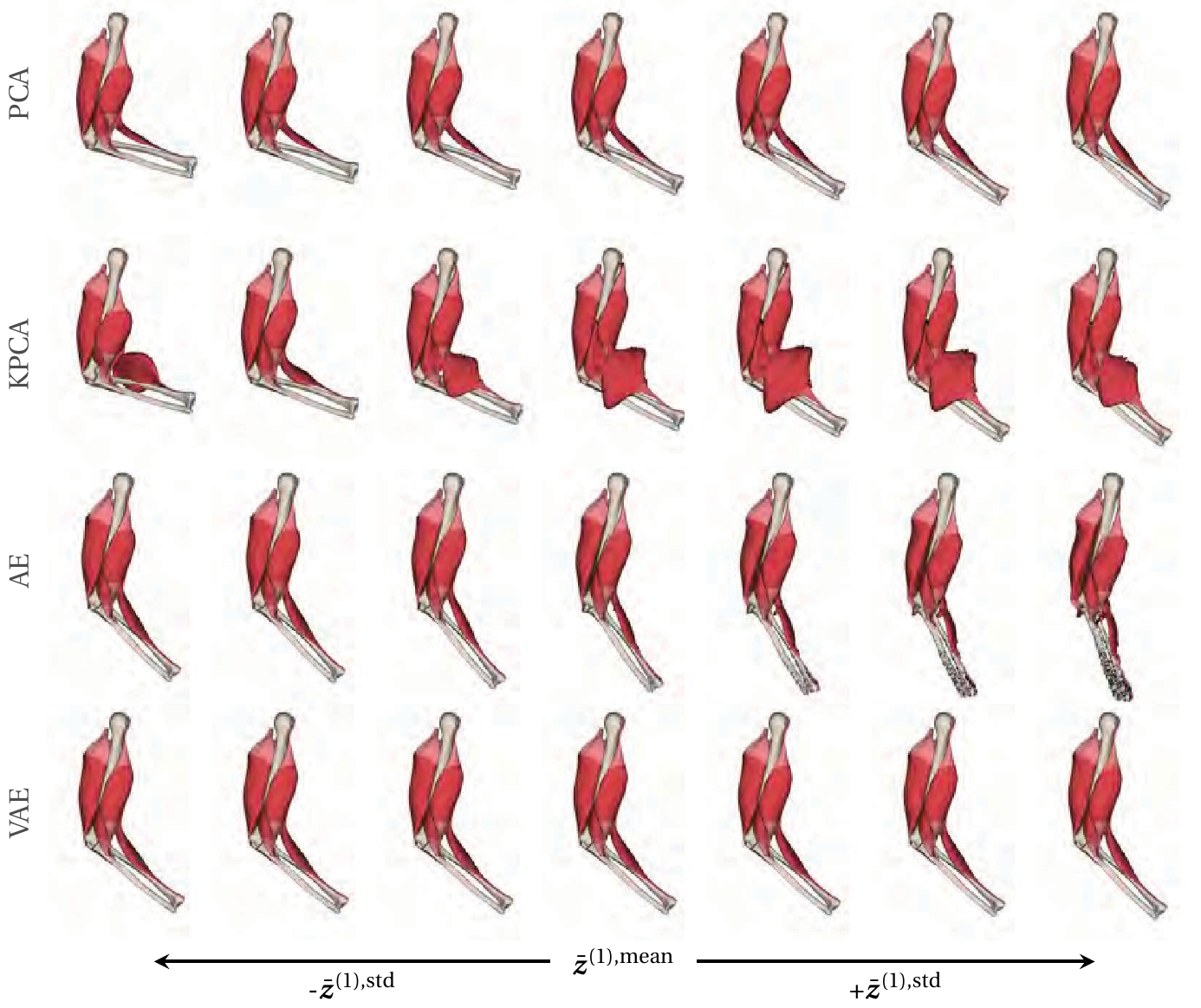}
	\caption{Reconstructed arms for reduced coordinates sampled around the mean of the first entry for \PCA, \KPCA, \AE, and \VAE.}
	\label{fig: modes}
\end{figure*}

\paragraph*{Sampling the Reduced Space}
As first comparison of the reduction algorithms, we present the impact that variations in the reduced space have on the original space. Therefore, the first entry~$\bar{\stateScalar}^{(1)}$ of the reduced state is varied around its mean value~$\bar{\stateScalar}^{(1), \text{mean}}=\frac{1}{\nSnaps}\sum_{j=1}^{\nSnaps}\bar{\stateScalar}^{(1)}_{j}$ with~$ \state\in\dataset_{\text{red}}$ occurring in the training data set. It is sampled within the range of the standard variance~$\bar{\stateScalar}^{(1), \text{std}}$. Please note that the mean and standard variance obviously differ for different reduction algorithms. For an isolated view all other entries~$\bar{\stateScalar}^{(2)},...,\bar{\stateScalar}^{(\redStateDim)}$ remain zero. The resulting reduced states are then transformed into a reconstructed arm~$\stateRec^{(\redIterator)}=\reconstruction([\bar{\stateScalar}^{(1)},0,...,0]^T)$. which is shown in \figref{fig: modes} for each reduction algorithm.

In the case of \PCA, this representation corresponds to the visualization of the first reduced basis vector. It can be observed that the first reduced basis vector of the \PCA is described by a motion during which the arm is raised. A similar arm raise can be observed for the \KPCA but simultaneously a non-physical muscle deformation occurs, especially in the brachioradialis. For the autoencoder, non-physical behavior can be observed especially in the lower-arm where it comes to bizarre deformations of the muscle and bone tissue.
Comparing this to the results of the \VAE, the benefits of regularizing the reduced space become apparent. The reconstructed arm of the \VAE exhibits such non-physical behavior to a much lesser extent and is mainly characterized by a slight lowering of the lower-arm.
It is not surprising that the arm raising occurs in all reduction algorithms as it is one of the most important motions. However, the individual modes from the \PCA are sorted by importance and orthogonal, which leads to the first reduced basis vector being an isolated motion of the most important movement without any other effect occurring. For the \AE, the modes are in general neither orthogonal nor sorted by importance and thus different effects can occur simultaneously.
Besides this more abstract representation of the impact of reduced coordinates, concrete performance measurements are presented in the following section. 

\subsection{Performance Measurements}
The fitted reduction~$\reduction$ and regression algorithms~$\regression$ are used in the online algorithm~\ref{alg: Online} to approximate the displacement~$\disps$ and stress~$\stress$ of the arm on the test data set, i.e., on unseen muscle activations~$\param$. The results are summarized in Table~\ref{tab:perf} where~$\sRec$ is used for assessing of the isolated reduction algorithms, $\sRegr$ for the isolated regression algorithm, and $\sApprox$ for an overall impression. Moreover, the overall simulations averaged mean $\hat{e}_{2}^{\text{mean}}=\frac{1}{\nSnaps}\sum_{j=1}^{\nSnaps}{e}_{2,j}^{\text{mean}}$ and maximum $\hat{e}_{2}^{\text{max}}=\frac{1}{\nSnaps}\sum_{j=1}^{\nSnaps}{e}_{2,j}^{\text{max}}$ errors provide a physical interpretable metric and the computational time $\compTime$ enables a comparison of the computation time. 
%\the\textwidth
\begin{table}
	\setlength{\tabcolsep}{3pt}
	\centering
	\caption{Performance measurements}
	\label{tab:perf}
	\npdecimalsign{.}
	\nprounddigits{3}
	\begin{tabular}[c]{c c c c c c c}
		\toprule
		&~$\sRecMean$ &~$\sRegrMean$ &~$\sApproxMean$ &~$\hat{e}_{2}^{\text{mean}}$ &~$\hat{e}_{2}^{\text{max}}$ &~$\compTime$\\
		\midrule
		disp~$\disps$: & (-) & (-) & (-) & (mm)& (mm) & (ms)\\
		\PCA+\GP 
			&~$\numprint{0.989821188209938}$ 
			&~$\numprint{0.888945587783079}$ 
			&~$\numprint{0.969857047243369}$ 
			&~$ \nprounddigits{2} \numprint{0.505577199319912}$  
			& \nprounddigits{2}$\numprint{2.633113071}$  
			& \nprounddigits{2}${\numprint{8.335357189178466}}$ \\
		\KPCA+\GP
			&~$\textbf{\numprint{0.995827916200426}}$ 
			&~$\numprint{0.888950028851718}$ 
			&~$\numprint{0.954053885850451}$ 
			& ~$\nprounddigits{2}\numprint{0.722225859166429}$    
			&\nprounddigits{2}~$\numprint{5.740967049}$ 
			& \nprounddigits{2}$\numprint{14.407105922698974}$\\
		\AE+\GP
			&~$\numprint{0.988485119445869}$ 
			&~$\textbf{\numprint{0.920377638016816}}$ 
			&~$\textbf{\numprint{0.971224313011645}}$ 
			& ~$\nprounddigits{2}\textbf{\numprint{0.446941636406031}}$  
			& \nprounddigits{2}$\textbf{\numprint{2.56624395}}$ 
			&\nprounddigits{2}~$\textbf{\numprint{3.6331920623779296}}$ \\
		\VAE+\GP
			&~$\numprint{0.980425837364559}$ 
			&~${\numprint{0.909296529557983}}$ 
			&~${\numprint{0.968356552427233}}$ 
			& ~$\nprounddigits{2}\numprint{0.476986062830071}$  
			& \nprounddigits{2}$\numprint{3.201416091}$ 
			&\nprounddigits{2}~${\numprint{3.679364204406738}}$ \\
		\midrule
		stress~$\stress$: & (-) & (-) &(-) & (MPa) & (MPa) & (ms)\\
		\PCA+\GP
			&~$\numprint{0.9361696615931971}$ 
			&~$\numprint{0.912054039035119}$  
			&~$\numprint{0.926578967385251}$ 
			&  \nprounddigits{3}$\numprint{0.0196834400112654}$ 
			& \nprounddigits{2}${\numprint{4.897621637}}$ 
			& \nprounddigits{2}${\numprint{12.442193031311034}}$ \\
		\KPCA+\GP
			&~$\textbf{\numprint{0.943587181388954}}$ 
			&~$\textbf{\numprint{0.912054183029457}}$ 
			&~$\numprint{0.932246116016579}$ 
			&  \nprounddigits{3}$\numprint{0.0153523964650394}$  
			& \nprounddigits{2}$
			\textbf{\numprint{4.729751012}}$ 
			& \nprounddigits{2}$\numprint{17.945446014404298}$ \\
		\AE+\GP
			&~$\numprint{0.935213184700853}$ 
			&~$\numprint{0.896057924491315}$ 
			&~$\textbf{\numprint{0.933701622416575}}$ 
			&  \nprounddigits{3}$\textbf{\numprint{0.0133256155166395}}$  
			& \nprounddigits{2}$\numprint{5.110950415}$ 
			& \nprounddigits{2}$\numprint{3.8864970207214355}$ \\
		\VAE+\GP
			&~$\numprint{0.937892728600554}$ 
			&~$\numprint{0.825103714194524}$ 
			&~${\numprint{0.933329908620833}}$ 
			&  \nprounddigits{3}$\numprint{0.0141488607794192}$  
			& \nprounddigits{2}${\numprint{4.80954532}}$ 
			& \nprounddigits{2}$\textbf{\numprint{3.609621524810793}}$
		%~$\speedup$\,(-)  & & & & & \\ 
%		\bottomrule
	\end{tabular} 
	\npnoround
\end{table}

% \begin{table}[h!]
% 	\begin{center}
% 	  \caption{Autogenerated table from .csv file.}
% 	  \label{table1}
% 	%   \pgfplotstabletypeset[
% 	% 	multicolumn names, % allows to have multicolumn names
% 	% 	col sep=comma, % the seperator in our .csv file
% 	% 	display columns/0/.style={
% 	% 	  column name=$Value 1$, % name of first column
% 	% 	  column type={S},string type},  % use siunitx for formatting
% 	% 	display columns/1/.style={
% 	% 	  column name=$Value 2$,
% 	% 	  column type={S},string type},
% 	% 	every head row/.style={
% 	% 	  before row={\toprule}, % have a rule at top
% 	% 	  after row={
% 	% 		  \si{\ampere} & \si{\volt}\\ % the units seperated by &
% 	% 		  \midrule} % rule under units
% 	% 		  },
% 	% 	  every last row/.style={after row=\bottomrule}, % rule at bottom
% 	%   ]{fig/results/boxplot/disp/PCA_overall_metrics_displacement.csv} % filename/path to file
% 	\end{center}
% \end{table}

It confirms that the stress data is more difficult to reconstruct, given that~$\sRec$ is consistently worse for stress than for displacement. In this context, it is noticeable that the \KPCA reaches the best reconstruction score~$\sRec$ and is accordingly most capable of identifying a low-dimensional representation of the data.
Regarding the regression score~$\sRegr$, the \GP that is trained on the data reduced by \PCA and \KPCA performs very similarly. This suggests that the identified reduced coordinates are similarly learnable.
Overall, the combination of an \AE and \GP yields the best results regarding the approximation score~$\sApprox$ as well as most of the physical error metrics. The variational autoencoder achieves worse results for~$\sRegr$ and consequently its reduced coordinates seem to be more difficult to approximate. However, the \VAE is at the same time more robust against these errors in the reduced space and almost reaches the same quality as the original \AE but with a more interpretable and regularized reduced space.

Regarding the computational time, the surrogate models using autoencoders are fastest followed by \PCA, whereas \KPCA is outperformed by far. The different computational times are only partially meaningful as the reduction algorithms are implemented using different software packages. While scikit-learn~\cite{scikit} is used for \PCA and \KPCA, TensorFlow~\cite{tensorflow2015} is used for the autoencoders.
In an even comparison, \PCA should be faster than the autoencoders as it contains less parameters. 
Nevertheless, all surrogate models can predict the arm behavior within milliseconds and thus are suitable for real-time applications.
It should be noted, that the computational times presented in this work are the averaged times for the approximation off a single sample. When several samples are calculated at once the required time per sample decreases significantly.
\\

\begin{figure}
	\begin{subfigure}[t]{0.49\textwidth}
		\includegraphics{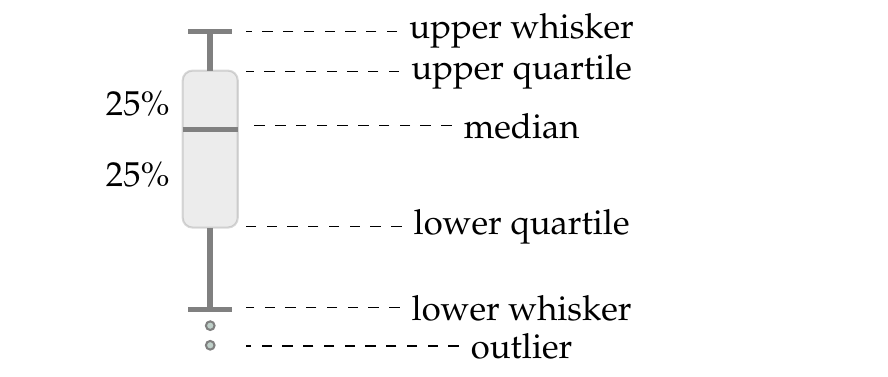}
		\caption{boxplot explanation}
		\label{fig: explanation boxplot}
	\end{subfigure}
	\begin{center}
		\includegraphics{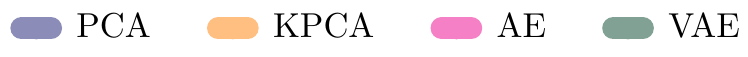}
	\end{center}
	\begin{subfigure}[t]{0.49\textwidth}
		\includegraphics[]{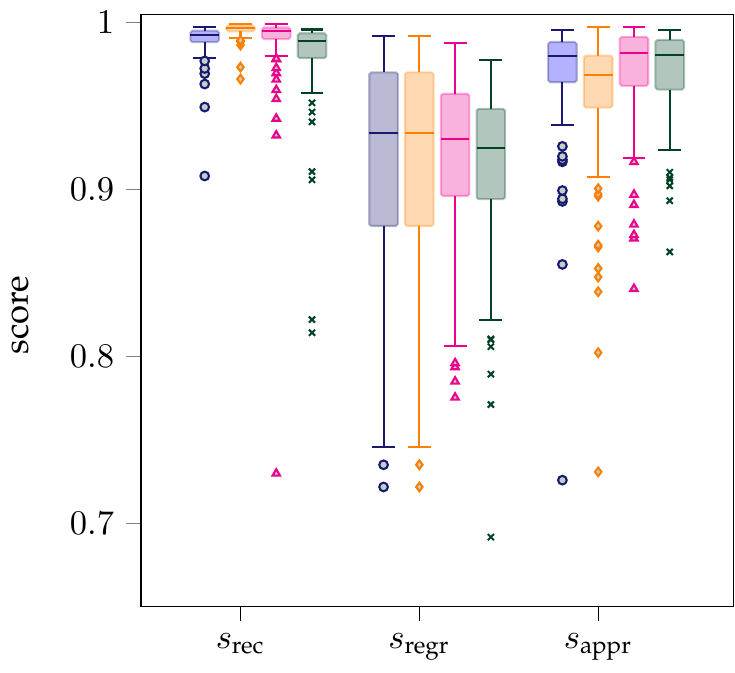}
		\caption{displacement}
		\label{fig: disp boxplot}
	\end{subfigure}
	\hfill
	\begin{subfigure}[t]{0.49\textwidth}
		\includegraphics[]{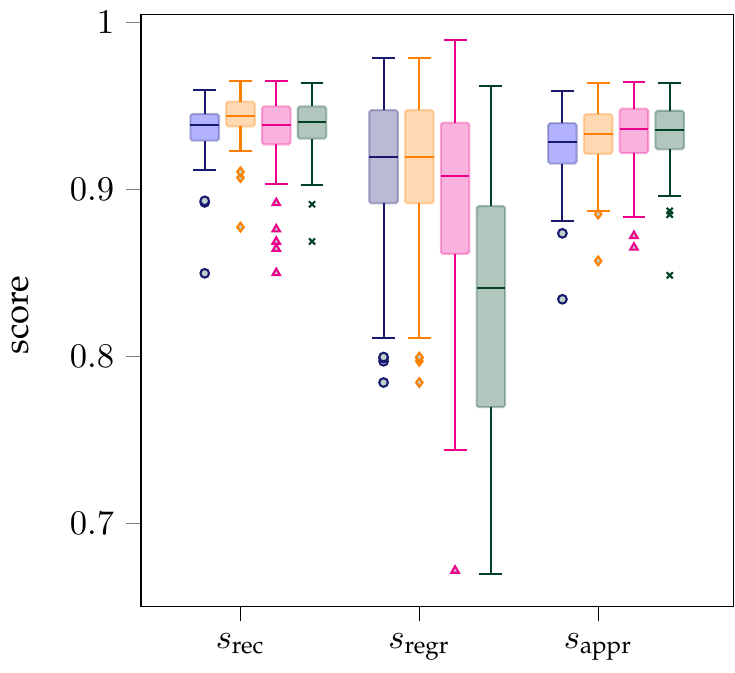}
		\caption{stress}
		\label{fig: stress boxplot}
	\end{subfigure}
	\caption{Performance scores achieved by surrogate models using different reduction algorithms for displacement and stress data.}
	\label{fig: boxplot scores}
\end{figure}

For a more detailed elaboration of the results, the distribution of the performance scores of all test simulations are considered in the following paragraph. On the one hand, this is intended to illustrate the relationships between the reduction, regression, and overall approximation performance and, on the other hand, to depict the differences between the individual reduction methods. 
For this purpose, the distribution of the relative performance scores of the different reduction algorithms are shown in \figref{fig: boxplot scores}. There a boxplot is used as representation and a visual explanation is given in \figref{fig: explanation boxplot}. It consists of a box which is divided into quartiles so that 25\% of the values are above the median value (upper quartile) and 25\% below (lower quartile). Furthermore, the box is extended by lines, so-called whiskers, which are chosen to have a maximum length of $1.5*IQR$. In this context, the interquartile range~$IQR$ is the distance between the upper and lower quartiles. Outliers not covered inside the range of the whiskers are visualized as individual points.

Considering the achieved results for the node displacements shown in \figref{fig: disp boxplot}, the isolated reconstruction score attests the \KPCA's ability to represent the data in a low-dimensional space without much loss of information. 
Furthermore, the reduced coordinates resulting from \PCA and \KPCA seem to be identically challenging to learn as the distribution of the regression score achieved by the \GP is almost identical. The overall approximation quality mainly suffers from the error induced by the regression as it deviates largely to the reconstruction score. Interestingly, the overall approximation score using the \KPCA is influenced to a larger extent as it is the case for the \PCA and accordingly seems to be less robust against disturbances in the reduced space with the chosen hyper parameters. For the autoencoders, a better regression score is achieved than for \PCA and \KPCA and they perform comparably similar.

Considering the scores achieved for the approximation of the arm's stress, as done in \figref{fig: stress boxplot}, the impression changes. The reduced coordinates obtained via \PCA and \KPCA seem to be easier to learn than for the autoencoders and thus they result in better regression scores but at the same time they still contribute to the overall approximation. Quite the opposite is shown by the results for the autoencoders, which reveal that the reduced coordinates are difficult to learn but errors in the reduced space propagate to a much lesser extent to the overall approximation. Consequently, the approximation score~$\sApprox$ stays close to the reconstruction score~$\sRec$.
This is especially evident for the \VAE, for which the worst regression score but the best approximation is achieved. Hence, it seems to be very robust against errors introduced by the regression what may be explained by the regularization of the reduced space.
Although the \AE does in most cases not place first in either of the reconstruction and regression categories, it still achieves the slightly better overall approximation for displacements as well as stress. The reason for this is that the errors of the individual categories do not affect the overall result to such an extent as is the case for \KPCA which is sensitive to errors in the regression. 

Comparing the displacement and stress data, two things are noteworthy. As expected the stress data poses the greater challenge for the reconstruction. Hence, the overall approximation is mainly limited by the reconstruction whereas the regression is the decisive factor for the displacements. At the same time, however, it must be noted that although worse results are achieved for the stress on average, there a less outliers than it is the case for the displacements. 
In addition to the distribution of the scores, the interested reader can view the performance scores for all individual test simulation in appendix \ref{secA1} in \figref{fig: scores} for a more detailed view.
% While the approximation of the displacements obtained with \KPCA yields the overall worst results, its satisfactory approximation score for the stress is relativized by outliers.
\\

\begin{figure}[ht!]
	\begin{center}
		\includegraphics{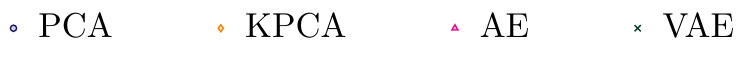}
	\end{center}
	\begin{subfigure}[t]{0.48\textwidth}
		\includegraphics[]{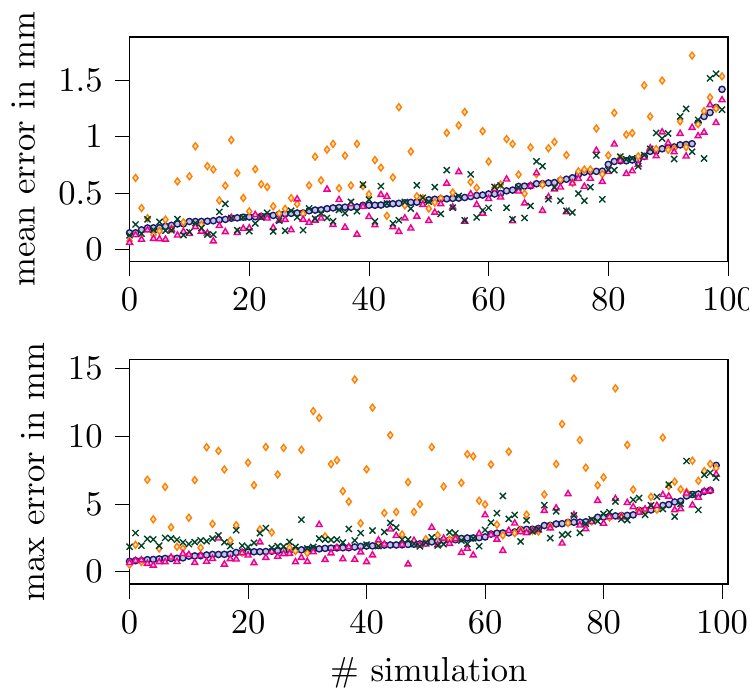}
		\caption{Mean and maximum node distance of displacement data.}
		\label{fig: disp errors}
	\end{subfigure}
	\hfill
	\begin{subfigure}[t]{0.48\textwidth}
		\includegraphics[]{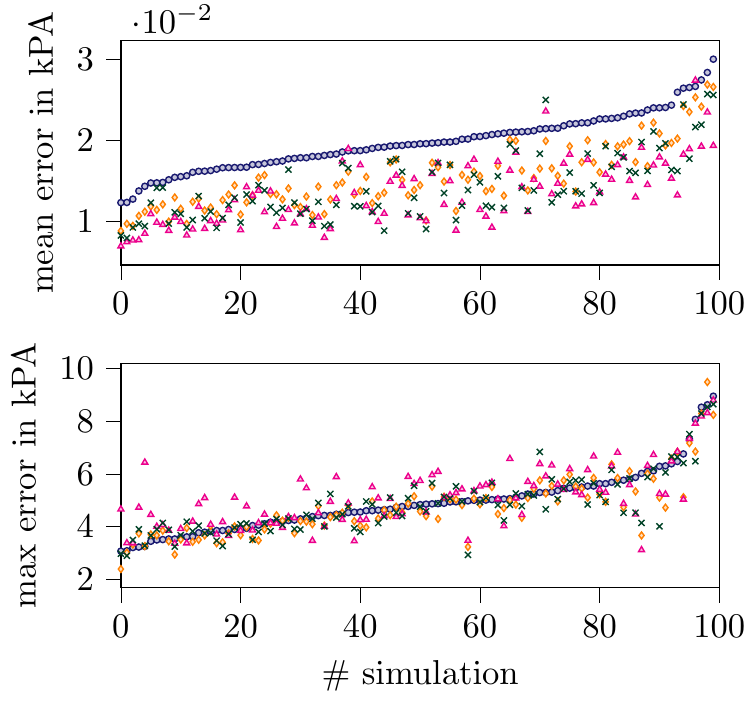}
		\caption{Mean and maximum error of stress data.}
		\label{fig: stress errors}
	\end{subfigure}
	\caption{Mean and maximum errors for displacements (a) and stress (b). The plots are sorted by the error made by the surrogate using \PCA.}
\end{figure}

Besides the evaluation criteria already presented, physical errors are considered in the following in order to contextualize them in real terms. In \figref{fig: disp errors}, the mean~\eqref{eq: e2 mean} and the maximum reached Euclidean node distance~\eqref{eq: e2 max} are shown for the three reduction algorithms. 
All of them manage to stay below a mean node distance of \SI{1.75}{\milli\metre} for all 100 test simulations. \PCA, \AE, and \VAE are within a similar range but \KPCA performs worst. The same applies for the maximum node distance. The rather large values for the \KPCA compared to the moderate mean values suggests significant outliers in the approximation of single nodes or areas. 

\PCA is not able to provide a low mean error for the stress data as visualized in \figref{fig: stress errors}. Here, it can be seen that all nonlinear reduction methods have a significant advantage. On the contrary, for the maximum error \PCA yield a similar accuracy and all algorithms are within a similar range. This suggests that \PCA is good at capturing local areas of high peaks, but does not perform that well for the entire model. 
\\

\begin{figure*}
		\centering
		\includegraphics[]{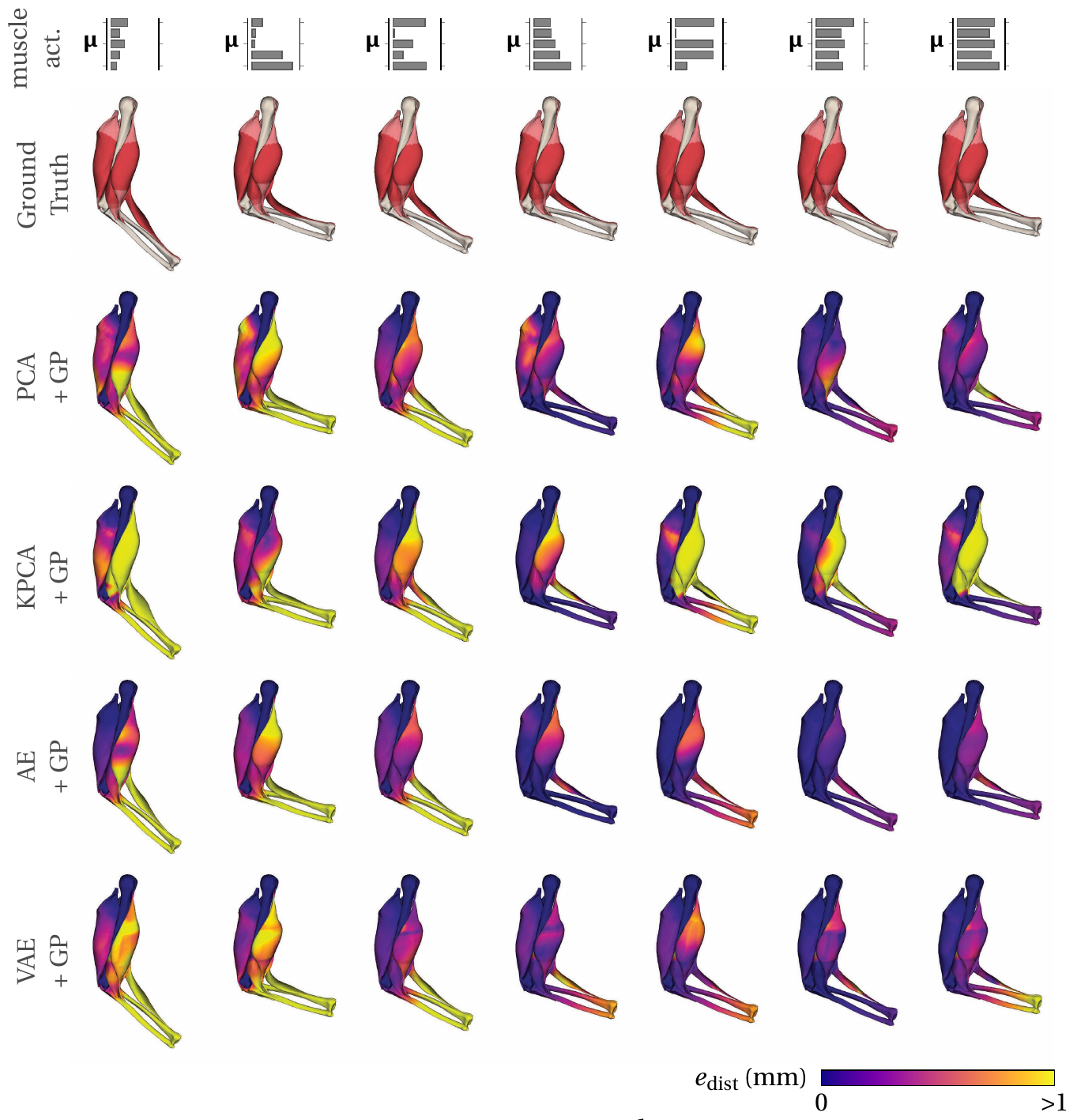}
		% \caption{Displacements}
		\caption{Visualization of the displacements of test simulations comparing the reference results (top) with their approximations using \PCA, \KPCA, \AE, and \VAE (from top to bottom). The approximations are colored with respect to the corresponding error of the individual nodes.}
	\label{fig: anim disp}
	% \label{fig: anim}
\end{figure*}

In addition to the numeric criteria presented so far, a visual assessment is given in the following to provide information about the most challenging parts of the arm model. 
A random set of seven simulations is picked from the test set and shown in \figref{fig: anim disp}. The finite element simulation results, i.e. the supposed ground truth, is shown at the top with the corresponding muscle activation of that simulation visualized as bar plot above. The results achieved by the three surrogate models are placed below with the nodes colored by their individual node distance~$\eTwo$.   
% \begin{figure*}
% 		\includegraphics[]{fig/results/animation/node_displacements_rel}
% 		% \caption{Displacements}
% 	\caption{Visualization of a test simulation comparing the reference results (left) with their approximations using \PCA, \KPCA, and \AE (from left to right). The approximations are colored with respect to the corresponding error of the individual nodes. \todo{Add Reconstructions without regression as well? Adjust colorbars so it is clear that they don't represent the complete range}}
% 	\label{fig: anim disp rel}
% 	% \label{fig: anim}
% \end{figure*}
It becomes particularly apparent which arm regions are most difficult to reproduce. This includes the complete forearm including the bones radius and ulna as well as the brachioradialis muscle on the one hand, and the biceps brachii on the other hand. In particular, the forearm is the part with the largest displacement so it is not surprising to find him in the worst approximated parts. \\
With the individual surrogate models, it is particularly noticeable that the one which relies on \PCA and \KPCA seem to have larger problems reproducing the triceps bachii. Nevertheless, the error in this region is low for all surrogates. The combination of \KPCA and \GP, furthermore, leads to large errors in the human arm for some muscle activations. For example, the complete biceps brachii is worse approximated in the last three visualized simulations. In contrast to the motion of the arm, the areas of high loading/large stress occur at different points for the stress.
%%%%%%%%%%%%%%%%%%%%%%%%%%%%%%%%%%%%%%%%%%%%%
%  	stress errors

\begin{figure*}
	\centering
	\includegraphics[]{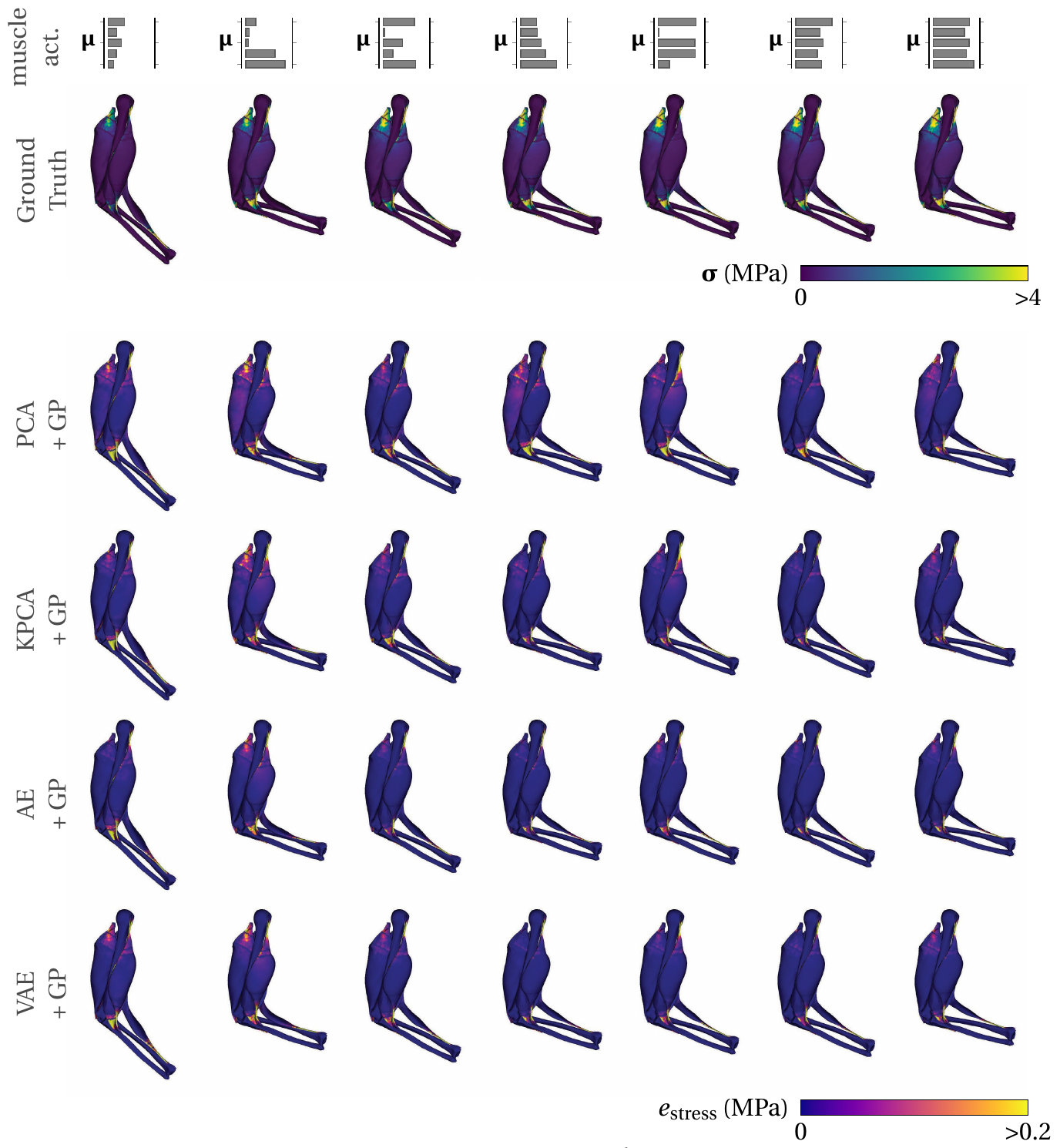}
	\caption{Visualization of the stress of test simulations comparing the reference results (top) with their approximations using \PCA, \KPCA, \AE, and \VAE (from top to bottom). The reference is colored with respect to the stress occurring in the individual node, while the approximations are colored with respect to the corresponding error.}
	\label{fig: anim stress}
\end{figure*}

As expected, the highest stress occurs primarily in the tendons, while only minor stress is found in the muscles as visualized in \figref{fig: anim stress} where the stress distribution obtained from the \FE model are presented at the top. Below of the reference solution, the approximation of the surrogate models colored by the respective errors are plotted. The largest discrepancies therefore unsurprisingly occur in the regions of highest stress. However, it is remarkable that the \PCA-based surrogate has larger areas of notable errors than the other surrogate models.
% \begin{figure*}
% 	\includegraphics[]{fig/results/animation/node_stress_rel}
% 	\caption{Visualization of a test simulation comparing the reference results (left) with their approximations using \PCA, \KPCA, and \AE (from left to right). The approximations are colored with respect to the corresponding error of the individual nodes. \todo{Add Reconstructions without regression as well? Adjust colorbars so it is clear that they don't represent the complete range}}
% 	\label{fig: anim stress rel}
% \end{figure*}

\subsection{Discussion}\label{sec:discussion}
The presented results demonstrate that the used methodology can create versatile surrogate models for a high-dimensional human arm model in a data-poor regime even for variables that are difficult to capture. The modular \LORE algorithm can be used to approximate stress as well as displacements. 
The surrogate models enable evaluation in real-time scenarios as they possess a very low computational time in the range of milliseconds. It is also possible to use them on resource-constrained hardware due to the savings in computational complexity. 

In comparing the different methods for the dimensionality reduction of the \FE data exciting differences are revealed. 
As expected \PCA performs well for displacements and is the simplest in its application. 
The results for the \KPCA, on the other hand, are contradictory. On the one hand, no other reduction algorithm reconstructed the data that well from the reduced/latent space. On the other hand, it is more sensitive to errors in the reduced/latent space introduced by the regression algorithm so that the overall approximation suffers significantly. There may be other kernel functions and hyperparameters with which those hurdles can be surmounted but none of the kernel function and hyperparameter combinations tested during the grid search yielded better results. 
The autoencoder delivers an overall satisfactory result except for the more expensive training phase. Although a fully connected \AE was used, contrary to expectations, we were able to succeed in a data-poor regime. Variational autoencoder have proven to be almost similar performant as the classical autoencoder but with a more interpretable reduced/latent space.

Even though slightly better results were obtained with the nonlinear reduction methods than with the linear PCA, the differences are marginal for the displacements. For the stress, on the contrary, it could be shown that \PCA was outperformed considering the mean error. However, the nonlinear methods did not achieve the significant advantage over the linear methods expected for the stress data. This could be due, among other things, to the fact that the autoencoders were trained with relatively little data.

Overall, it can be said that the \LORE algorithm is suitable for approximating high-dimensional quantities that are difficult to represent. Depending on the intended use, the individual components must be carefully selected. The Gaussian process regression was successfully combined with all reduction techniques. As one scope of this work was to compare the reduction methods, the \GP was not optimized for the individual combinations and consequently introduced a non neglectable error to the overall approximation. With further hyperparameter-tuning the overall error could be further decreased. Moreover, even though \GPs worked well in previous literature, other regression methods such as neural networks have also shown satisfying results. 
For the dimensionality reduction, on the contrary, the results hardly depend on the underlying problem. While all algorithms worked well for data like displacements they introduced a larger error when it comes to the reconstruction of stress data. 
The best results for stress were still achieved by the nonlinear methods, i.e., \KPCA, \AE and \VAE. However, since the first method has been found to be prone to outliers, the latter is generally preferable.

Besides the approximation quality measures, soft factors can also play a role in the choice of the reduction algorithm. Differences can be found, especially during the offline phase but also regarding the consumption of resources. While \PCA usually works right away, \KPCA \AE, and \VAE require a more extensive hyperparameter tuning including the choice of kernel function for the \KPCA and the network architecture for the \AE and \VAE. The latter also require a computational expensive training phase while the former consumes the most memory for evaluation.

\section{Conclusion}\label{sec:conclusion}

While real-time evaluation of complex structural dynamical problems has not been feasible for a long time, it is possible with today's resources and tools. Especially machine learning methods can serve as extremely fast enabler for surrogate models. With the combination of dimensionality reduction and a Gaussian process regressor we were able to create accurate yet real-time capable surrogate models for a complex human arm. In this process, not only often considered physical quantities like displacements but also often neglected stress values have been approximated. The latter turned out to be more challenging to be represented in a low-dimension without loosing to much information. Accordingly, nonlinear reduction techniques in the form of kernel principal component analysis, autoencoder, and variational autoencoder were implemented besides the linear reduction in the form of the principal component analysis. 
All methods were able to process the displacements satisfactorily. However, for the challenging reduction of the stress data the nonlinear reduction methods only slightly outperformed the linear one which may be owed to the comparatively small data basis used. Nonetheless, the stress behavior of the complex arm model could be satisfyingly and quickly approximated by the surrogate models. As soon as disturbances in the reduced coordinates occur, differences between them become apparent. This means errors induced by the Gaussian process approximating the reduced states can lead to non neglectable errors in the overall approximation, especially for the \KPCA-based surrogates. Overall, the autoencoder-based surrogate yielded the best results while the one using \PCA performs well albeit its simple application.

We summarize the the effects of the various reduction techniques:
\begin{enumerate}[label=\roman*)]
	\item {Principal Component Analysis}: \\ \PCA Captivates with its simplicity and is computationally the most favorable method in the offline ans online phase as neither expensive hyperparameter tuning nor an expensive training phase is necessary. However, \PCA can reach its limits when linear reduction, i.e., a linear combination of reduced basis vectors, is a too great restriction. 
	\item {Kernel Principal Component Analysis}: \\ \KPCA yields the best reconstruction from coordinates transformed into the reduced space. However, it is more sensitive to errors in the reduced space and thus leads to larger overall errors. No hyperparameter are found which were able to reconstruct well and robust. The offline phase including the hyper parameter tuning is expensive and the memory consumption non neglectable.
	\item {Autoencoder}: \AEs provided the best overall approximation and still are fast. These advantages are bought by means of an expensive training phase. Furthermore, they require a lot of data which may limits their applicability for certain problems. 
	\item {Variational Autoencoder}: \VAEs regularize the reduced space leading to more interpretable reduced coordinates without losing much reconstruction capability compared to the normal autoencoder. Otherwise they behave similarly.
\end{enumerate}
\backmatter

% \bmhead{Supplementary information}
 
\section*{Declarations}
\section*{Competing interests}
The authors have no relevant financial or non-financial interests to disclose. 

\section*{Availability of data}
The datasets generated during and/or analyzed during the current study are available in the DaRUS repository, see~\cite{darusArm}.
% bib commented out for fast compilation
% % \bibliography{sn-bibliography}% common bib file
% \bibliographystyle{bst/sn-basic}
% \bibliography{sn-bibliography.bib}

% \clearpage
\newpage
\begin{appendices}
\section{Constitutive parameters}
% \DRcomment{these are the parameters for the biceps. I will fill in the rest tomorrow.}
\begin{table}[h]
	\centering
	\caption{\textbf{Model parameters.} Model parameters for the muscle-tendon-complex of the arm model. For $c_i$ superscripts M and T differentiate between muscle and tendon.}
	\begin{tabular}{c c}%{|c|c|c|}
		\hline
		\textbf{Parameter} & \textbf{Value}\\% & \textbf{Source}\\
		\hline
		$c_1$ & $1.00\cdot 10^{-1}$ MPa \\%& \multirow{2}{2cm}{\citeauthor{hawkins1994comprehensive}}\\
		$c_2$ & $2.00\cdot 10^{-2}$ MPa \\%& \\
		\hline
		$c^M_3$ & $1.00\cdot 10^{-6}$ MPa \\%& \multirow{2}{2cm}{\citeauthor{zheng1999objective}}\\
		$c^M_4$ & 35.00 \\%& \\
		\hline
		$c^T_3$ & 6.50 MPa \\%& \multirow{2}{2cm}{\citeauthor{weiss2001computational}}\\
		$c^T_4$ & 25.00 \\%& \\
		\hline
		$\Delta \Gamma_{asc}$ & 0.15 \\%& \multirow{6}{2cm}{adapted from \citeauthor{gunther2007high}}\\
		$\Delta \Gamma_{desc}$ & 0.16 \\%& \\
		$\nu_{asc}$ & 2.00 \\%& \\
		$\nu_{desc}$ & 6.00 \\%& \\
		$\lambda^{opt}_f$ & 1.30 \\%& \\
		$\sigma_{max}$ & 2.00 MPa \\%& \\
		\hline
	\end{tabular}
\label{tab:params}
\end{table}

%Anconeus: c1=0.1, c2=0.02, c3=1e-6 (M), c4=35 (M), c3=6.5 (T), c4=25.0 (T), lamnda_opt=1.3, W_asc=0.15, v_asc=2.0, W_desc=0.16, v_desc=6.0
\newpage
\section{Simulation Results}\label{secA1}
\begin{figure}[h!]
	\begin{center}
		\includegraphics{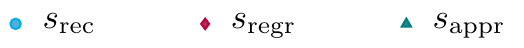}
	\end{center}
	\begin{subfigure}[t]{0.475\textwidth}
		\includegraphics[width=\linewidth]{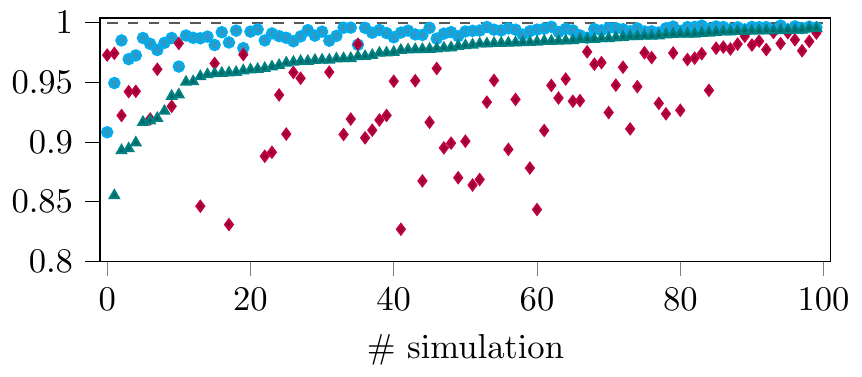}
		\caption{displacement: \PCA+ \GP}
		\label{fig: disp scores PCA}
	\end{subfigure}
	\hfill
	\begin{subfigure}[t]{0.475\textwidth}
		\includegraphics[width=\linewidth]{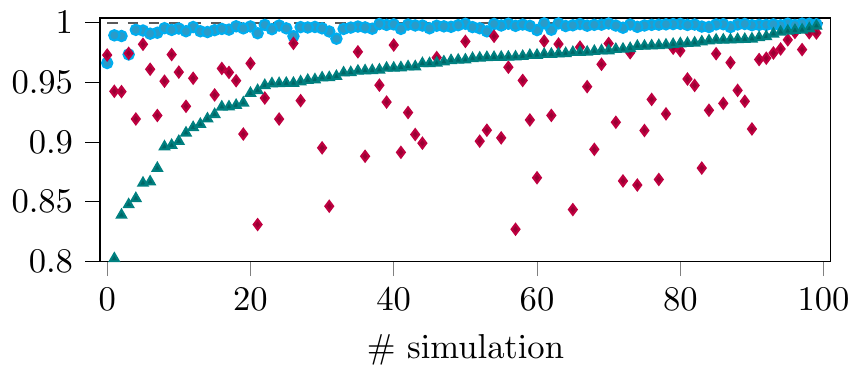}
		\caption{displacement: \KPCA+ \GP}
		\label{fig: disp scores KPCA}
	\end{subfigure}
	\hfill
	\begin{subfigure}[t]{0.475\textwidth}
		\includegraphics[width=\linewidth]{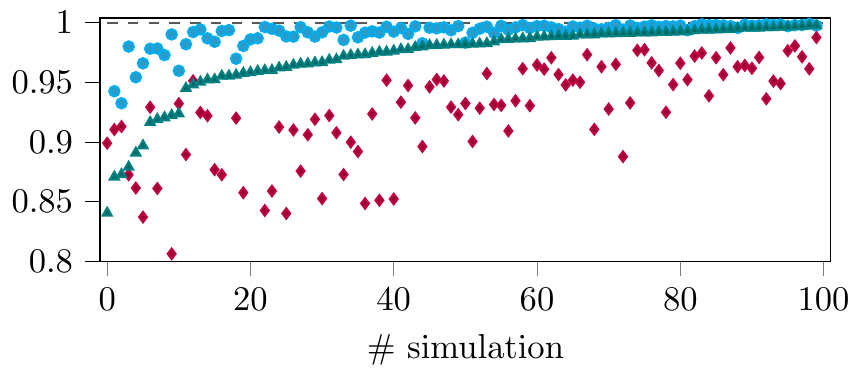}
		\caption{displacement: \AE+ \GP}
		\label{fig: disp scores AE}
	\end{subfigure} 
	\hfill
	\begin{subfigure}[t]{0.475\textwidth}
		\includegraphics[width=\linewidth]{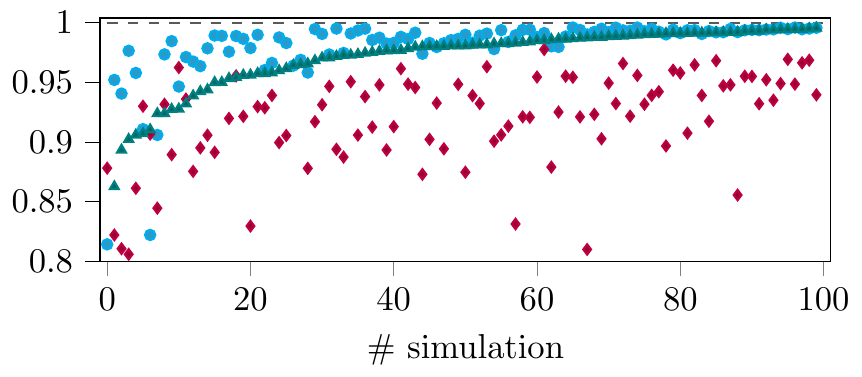}
		\caption{displacement: \VAE+ \GP}
		\label{fig: disp scores VAE}
	\end{subfigure} 
	\caption{Performance scores of displacement data for \PCA, \KPCA, \AE, and \VAE. The individual plots are sorted by the achieved approximation score~$\sRec$.}
\end{figure}
\newpage
~\newline ~\newline
\begin{figure}[b!]
	\hfill
	\begin{center}
		\includegraphics{fig/results/legend_score.pdf}
	\end{center}
	\hfill
	\begin{subfigure}[t]{0.475\textwidth}
		\includegraphics[width=\linewidth]{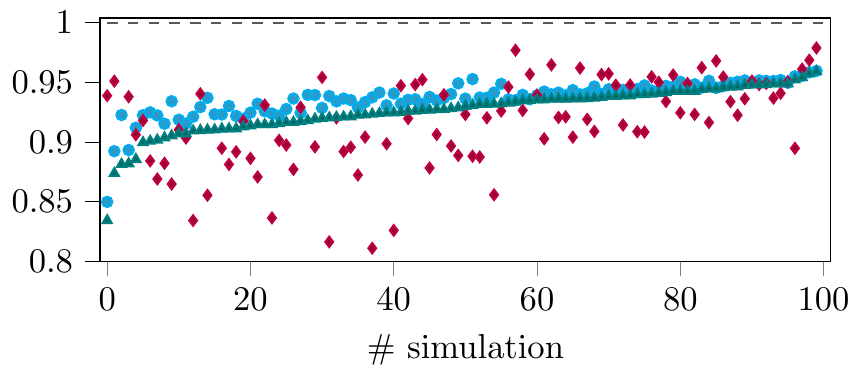}
		\caption{stress: \PCA+ \GP}
		\label{fig: stress scores PCA}
	\end{subfigure}
	\hfill
	\begin{subfigure}[t]{0.475\textwidth}
		\includegraphics[width=\linewidth]{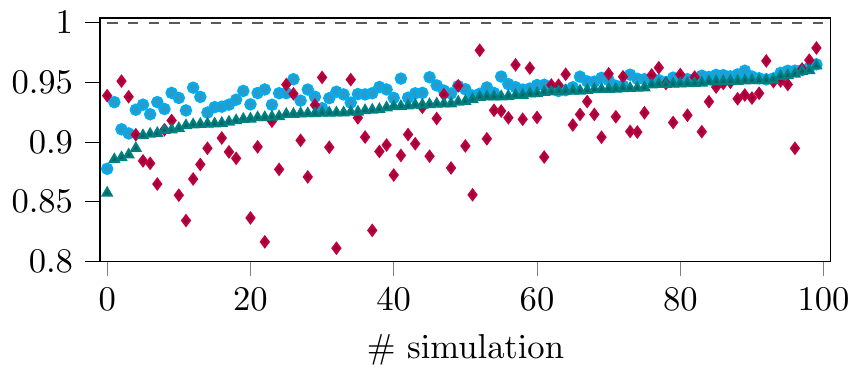}
		\caption{stress: \KPCA+ \GP}
		\label{fig: stress scores KPCA}
	\end{subfigure}
	\hfill
	\begin{subfigure}[t]{0.475\textwidth}
		\includegraphics[width=\linewidth]{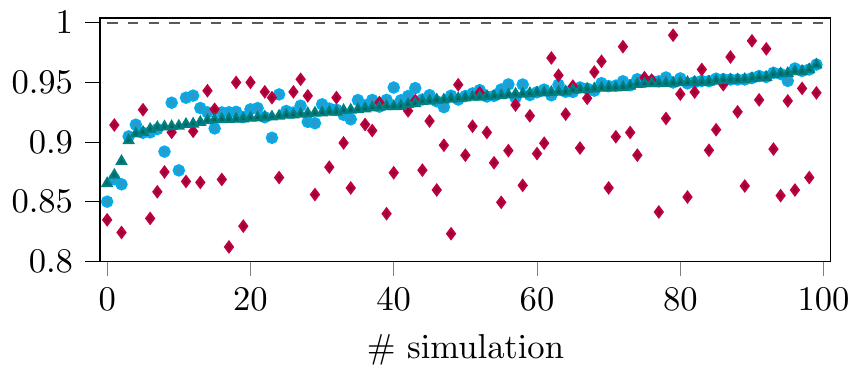}
		\caption{stress: \AE+ \GP}
		\label{fig: stress scores AE}
	\end{subfigure}
	\hfill
	\begin{subfigure}[t]{0.475\textwidth}
		\includegraphics[width=\linewidth]{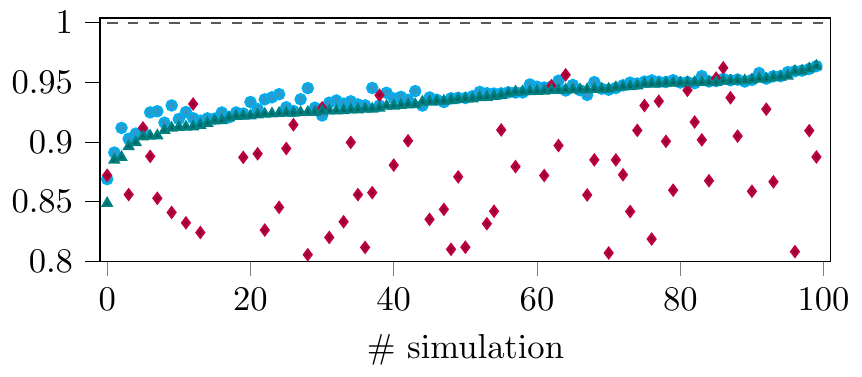}
		\caption{stress: \VAE+ \GP}
		\label{fig: stress scores VAE}
	\end{subfigure}
	\caption{Performance scores of stress data for \PCA, \KPCA, \AE, and \VAE. The individual plots are sorted by the achieved approximation score~$\sRec$.}
	\label{fig: scores}
\end{figure}

\end{appendices}
%%===========================================================================================%%
%% If you are submitting to one of the Nature Portfolio journals, using the eJP submission   %%
%% system, please include the references within the manuscript file itself. You may do this  %%
%% by copying the reference list from your .bbl file, paste it into the main manuscript .tex %%
%% file, and delete the associated \verb+\bibliography+ commands.                            %%
%%===========================================================================================%%
% \bibliography{/Users/jonaskneifl/Develop/12_itm_literatur/ITM_Literatur.bib}
% \bibliography{KneiflRosinRoehrleFehr}
% \bibliography{/Users/jonaskneifl/Develop/12_itm_literatur/ITM_Literatur.bib}%
%% if required, the content of .bbl file can be included here once bbl is generated
%%\input sn-article.bbl
% \clearpage
\bibliographystyle{bst/sn-basic}
\bibliography{sn-bibliography.bib}

\end{document}